\pdfoutput=1

\documentclass[11pt]{article}

\usepackage[final]{acl}

\usepackage{times}
\usepackage{latexsym}

\usepackage[T1]{fontenc}

\usepackage[utf8]{inputenc}

\usepackage{microtype}

\usepackage{inconsolata}

\usepackage{graphicx}

\usepackage{amsmath}
\usepackage{amsthm}
\usepackage{amssymb}
\usepackage{booktabs}
\usepackage{multicol}
\usepackage{multirow}
\usepackage{algorithm}
\usepackage{dsfont}
\usepackage{array}
\usepackage{subcaption} 
\usepackage{caption}
\usepackage{setspace}

\title{A Dual-Perspective NLG Meta-Evaluation Framework with Automatic Benchmark and Better Interpretability}

\author{Xinyu Hu,  Mingqi Gao, Li Lin, Zhenghan Yu, Xiaojun Wan\\
Wangxuan Institute of Computer Technology, Peking University\\
\texttt{\{huxinyu,gaomingqi,wanxiaojun\}@pku.edu.cn} \\
\texttt{\{efsotr\_l,zhenghanyu\}@stu.pku.edu.cn}}

\begin{document}
\maketitle

\begin{abstract}
In NLG meta-evaluation, evaluation metrics are typically assessed based on their consistency with humans. However, we identify some limitations in traditional NLG meta-evaluation approaches, such as issues in handling human ratings and ambiguous selections of correlation measures, which undermine the effectiveness of meta-evaluation. In this work, we propose a dual-perspective NLG meta-evaluation framework that focuses on different evaluation capabilities, thereby providing better interpretability. In addition, we introduce a method of automatically constructing the corresponding benchmarks without requiring new human annotations. Furthermore, we conduct experiments with 16 representative LLMs as the evaluators based on our proposed framework, comprehensively analyzing their evaluation performance from different perspectives.
\end{abstract}

\section{Introduction}

Natural Language Generation (NLG) has long been crucial in natural language processing research, encompassing common tasks such as text summarization and dialogue response generation. Unlike other tasks, such as question answering and mathematical reasoning, which have a single standard answer, making it convenient to evaluate the model output by direct matching, NLG tasks are inherently open-ended and lack a unique correct output. Therefore, they require more robust and flexible evaluation metrics. Traditional metrics primarily include string matching-based methods, such as BLEU \citep{papineni2002bleu} and ROUGE \citep{lin2004rouge}, as well as semantic representation-based methods like BERTScore \citep{zhangbertscore} and BARTScore \citep{yuan2021bartscore}. However, with the advancements of language models, these metrics have gradually been found insufficient to meet requirements. Moreover, some studies \citep{sai2021perturbation, he2023blind} have highlighted their deficiencies in robustness.

Recently, large language models (LLMs) such as GPT-4 \citep{achiam2023gpt} have demonstrated remarkable leaps in text comprehension and instruction-following capabilities. As a result, many studies \citep{wang2023chatgpt,chiang2023can} have shifted their focus to leveraging carefully designed instructions to directly prompt LLMs for simulating human evaluation. Additionally, to further enhance the evaluation capabilities of LLMs, some works have constructed training data specifically for evaluation scenarios to fine-tune LLMs. This LLM-based evaluation paradigm, known as LLM-as-a-Judge, has been recognized as well-performing in evaluating common NLG tasks \citep{wang2023chatgpt, liu2023g}. Moreover, it has already been widely adopted in many automated evaluation scenarios, such as the well-known AlpacaEval \citep{alpaca_eval}.

Nevertheless, we argue that traditional NLG meta-evaluation used to assess the performance of metrics has certain limitations, preventing a comprehensive understanding of their evaluation capabilities. Firstly, when measuring the consistency between metrics and human evaluations, multiple human ratings for each sample are typically averaged directly. However, human ratings are not continuous values and may be inconsistent among different annotators, and the quality intervals between adjacent ratings are not uniform \citep{sullivan2013analyzing}. Therefore, this seemingly intuitive averaging aggregation is not such reasonable. In addition, the choice of correlation measures commonly used in traditional meta-evaluation can significantly impact the performance assessment of different metrics \citep{perrella2024guardians, gao2024analyzing}. It is unclear which correlation measure should be selected in different NLG evaluation scenarios. Furthermore, many widely used NLG evaluation benchmarks suffer from issues including outdated generation systems and potential data contamination, further undermining the reliability of meta-evaluation. More detailed discussions of these limitations are included in Section \ref{sec:3}.

To address these issues, we propose a new dual-perspective NLG meta-evaluation framework. Specifically, the meta-evaluation from the global perspective formulates evaluation as an ordinal classification task, assessing the ability of metrics to judge the coarse-grained quality ratings of targets. On the other hand, the local-perspective meta-evaluation formulates evaluation as an adjacent pairwise comparison task, assessing the metrics in distinguishing between targets with fine-grained quality differences. These two perspectives focus on different evaluation capabilities, forming a more interpretable NLG meta-evaluation approach compared to the traditional meta-evaluation.

Furthermore, we introduce a method of automatic benchmark construction to better implement our proposed dual-perspective meta-evaluation. Our new benchmarks can be automatically constructed based on any existing NLG evaluation benchmark with new content, which avoids the cost of additional human annotations and potential data contamination. To comprehensively assess the evaluation capabilities of emerging LLM-as-a-Judge, we further experiment with 16 representative LLMs, including both the general-purpose and specifically fine-tuned ones, based on our new NLG meta-evaluation framework. The results indicate LLMs have their respective proficiency in different evaluation capabilities, demonstrating the effectiveness and necessity of our framework.

Overall, our main contributions are as follows:
\begin{enumerate}
    \item We reveal several limitations in traditional NLG meta-evaluation paradigms and benchmarks and propose a new dual-perspective meta-evaluation framework, which emphasizes different evaluation capabilities while offering better interpretability.
    \item We introduce a method of automatically constructing benchmarks for our proposed dual-perspective NLG meta-evaluation, which avoids the high cost of new human evaluations and the risk of data contamination.
    \item We conduct extensive experiments with 16 representative LLMs, analyzing their respective proficiency in different evaluation capabilities. Our code and data are released in \href{https://github.com/PKU-ONELab/NLG-DualEval}{https://github.com/PKU-ONELab/NLG-DualEval} to facilitate related research.
\end{enumerate}

\section{Background}

In natural language generation (NLG) evaluation, we are focused on the quality of outputs of a specified NLG task. Aside from human evaluation, which is usually considered the gold standard, various automatic metrics are applied to improve evaluation efficiency, such as BLEU and BERTScore, and the recently emerging LLM-as-a-Judge method. Furthermore, the performance analysis of these metrics is referred to as NLG meta-evaluation, which typically uses specific benchmarks to calculate the consistency between the evaluation results from humans and metrics. We present the definitions of key relevant concepts as follows.

\paragraph{Source} The input of the NLG task (e.g., the article in the text summarization). Typically, an evaluation benchmark will contain $n$ different sources $\mathcal{S} = \{s_i\}_{i=1}^n$.

\paragraph{Target} The output of the NLG task (e.g., the summary in the text summarization), which is usually generated for each source $s_i$ by each of the $m$ systems. These targets $\mathcal{T} = \{t_{ij}\}_{i=1,j=1}^{n,m}$ constitute the primary objects of meta-evaluation.

\paragraph{Evaluation Scale} The quantitative criterion for human evaluation, which is typically a five-point Likert scale (1-5), where rating 1 represents the worst quality and rating 5 represents the best. The scale is predefined in the evaluation guidelines, involving $l$ different ratings corresponding to $l$ distinguishable quality levels of the target during evaluation (e.g., $l$ equals 5 on a scale of 1-5).

\paragraph{Evaluation Result} The evaluation results of each target $t_{ij}$ from humans ($h_{ij}$) or metrics ($x_{ij}$). More specifically, the results from humans are usually presented in the form of discrete ratings based on the evaluation scale and may involve multiple annotators. Instead, the results from metrics are often deterministic and continuous scores, like BLEU. Although the method of LLM-as-a-Judge prompts LLMs to provide integer evaluation scores, they are often averaged across multiple samplings in practice for better performance \citep{liu2023g, chiang-lee-2023-closer}. It is approximately the weighted average of different scores based on the corresponding generation probabilities of LLMs, making the final results more fine-grained.

\paragraph{Aggregation Method} The method of converting multiple evaluation ratings from $a$ human annotators for each target into a single score that can be compared with that from the metric. The most common aggregation is to serve ratings as scores and then average them: $h_{ij} = \frac{1}{a} \Sigma_{k=1}^a h_{ij}^k$.

\paragraph{Consistency Measure} The method for calculating the evaluation consistency between final scores from humans and metrics. In traditional NLG meta-evaluation, the correlation method is commonly adopted, such as input-level Spearman correlation: $\frac{1}{n} \Sigma_{i=1}^n \rho (\{h_{ij}\}_{j=1}^m, \{x_{ij}\}_{j=1}^m)$.

\section{Dual-Perspective Meta-Evaluation}
\label{sec:3}

Recent studies \citep{perrella2024guardians, gao2024analyzing} have investigated the impact of different consistency measures on NLG meta-evaluation, but less attention has been paid to the aggregation method. The implementation of directly averaging ratings from multiple human evaluators has been followed by default in the previous related work. However, we have doubts about this seemingly self-evident approach from two main aspects. We take the widely-used benchmark, SummEval \citep{fabbri2021summeval}, as an example, where human evaluation is conducted using a 5-point Likert scale with three annotators.

\textbf{The targets with the same averaged rating do not necessarily have the same quality.} According to the studies \citep{sullivan2013analyzing, joshi2015likert} on the Likert scale, the quality intervals between adjacent ratings are not uniform. For example, the quality difference between ratings 4 and 3 may be greater than the difference between ratings 3 and 2. As a result, despite the same averaged rating, a target originally rated (2, 3, 4) by three annotators cannot be simply assumed to have the same quality as one rated (3, 3, 3). Moreover, targets like the former inherently reflect serious disagreement among human evaluators, which will undermine the reliability of meta-evaluation.

\textbf{The additional quality levels generated by averaging aggregation are not necessarily valid.} The averaged ratings with the 1-5 scale and three annotators can have up to 13 distinct values, leading to some new ratings besides the original five ratings, such as rating $\frac{4}{3}$. To verify whether these new ratings really reflect finer-grained quality levels beyond the original ratings, we conduct a human re-evaluation through pairwise comparison on SummEval. Specifically, we sample specific target pairs as the first test group, where the two averaged ratings involved in each pair are adjacent and belong to new ratings and original ratings, respectively, such as ratings $\frac{4}{3}$ and 1. As a contrast, we additionally sample the second test groups, where the two averaged ratings involved are adjacent original ratings, such as ratings 2 and 1. More experimental details are included in Appendix \ref{sec:A1}. The proportions of pairs with consistent judgments between our re-evaluations and the original averaged human ratings from SummEval are only 42\% for the first group, while 88\% for the second group. The results indicate that there are indeed definite quality differences among the original ratings, whereas this is not the case if adding the new ratings generated by averaging aggregation.

To avoid these issues, a straightforward aggregation method is to retain only targets with consistent ratings from multiple annotators. However, it introduces a new limitation of reducing the discriminative power of the benchmark. Specifically, the scale in traditional NLG meta-evaluation is usually coarse-grained, such as 1-5, which accounts for the limited human evaluation capacity to ensure the quality of annotations. Therefore, some targets with the same rating may actually have distinct qualities, e.g., if the more fine-grained evaluation scale (e.g., 1-10) and the more professional annotator are available. Unfortunately, such quality differences do not contribute to enhancing metric assessment under the traditional meta-evaluation paradigm. Instead, if a metric correctly distinguishes these differences, \textbf{it will be penalized rather than rewarded}, such as when calculating correlations. An example is provided in Appendix \ref{sec:example} for detailed illustration. To further address this limitation, we propose a new dual-perspective NLG meta-evaluation approach with different focuses on the assessment of evaluation capabilities.

\subsection{Global Perspective}

In global-perspective meta-evaluation, the focus is on the ability to judge the coarse-grained ratings of targets with various qualities rather than distinguishing quality differences within the same coarse-grained rating. It aims to avoid unreasonable penalization in traditional meta-evaluation mentioned before. Considering that the evaluation scale inherently possesses both categorical and ordinal properties, we treat the meta-evaluation as an ordinal classification task, where different ratings are treated as different categories. The ordinal classification and its dedicated metrics extend standard classification by considering the relationships between different categories, thereby enabling more discriminative assessment. For instance, while misclassifying a target with a true rating of 5 as rating 4 is incorrect, it is less severe than misclassifying it as rating 3. Finally, the performance of the evaluation metric is assessed as follows:
\begin{align*}
\mathrm{CEM}(\{h_{ij}\}_{i=1,j=1}^{n,m}, \{x_{ij}\}_{i=1,j=1}^{n,m}, l)
\end{align*} 
where $l$ represents the number of different coarse-grained ratings, and $\mathrm{CEM}$ represents the $\textit{Closeness Evaluation Measure}$, which is an effective ordinal classification metric proposed by \citet{amigo2020effectiveness}. More importantly, it does not require the assumption of equal intervals between adjacent categories, aligning with our previous discussions on the issues of the averaging aggregation method. More details and descriptions of $\mathrm{CEM}$ are included in Appendix \ref{sec:A2}.

\subsection{Local Perspective}
\label{sec:3.2}

On the other hand, in local-perspective meta-evaluation, the focus shifts to the ability to perceive fine-grained quality differences, thereby complementing the global perspective and rewarding those metrics with good discriminative ability. More specifically, given the targets with different qualities, where some of them may have the same coarse-grained rating, the concern here is whether they can be correctly ranked rather than whether they can be rated correctly. Furthermore, through a high-density target sequence for each source that contains significantly more than $l$ targets with different qualities, we can enable the more fine-grained meta-evaluation. The performance of the evaluation metric is assessed by comparing their evaluation scores for each pair of adjacent targets in the sequence sorted by target quality, which is the most challenging case. Given a target sequence $t_{i1}, t_{i2}, \cdots, t_{ik}$ for each source $s_i$ with the incremental target quality, the comparison accuracy is calculated as follows:
\begin{align*}
\frac{1}{n(k-1)}\sum_{i=1}^n \sum_{1\leq j < k}^{} \mathds{1}(x_{ij} < x_{i,j+1})
\end{align*}

\subsection{Preliminary Study on SummEval}
\label{sec:3.3}

We conduct a preliminary experiment on SummEval using our proposed meta-evaluation approaches, along with two traditional correlation measures, with detailed settings provided in Appendix \ref{sec:A3}. As shown in Table \ref{tab:preliminary_result}, different meta-evaluation approaches lead to differentiated performance rankings of evaluators, highlighting the complexity of meta-evaluation and the necessity of enhancing its interpretability. Although some work \citep{gao2024analyzing} has analyzed different correlation measures, their focus was primarily on evaluation stability in specific aspects. In contrast, our proposed meta-evaluation approach focuses on different evaluation capabilities, thereby offering more intuitive guidance for practical evaluation scenarios. For instance, global-perspective meta-evaluation is more suitable for scenarios requiring coarse-grained qualitative judgments, such as selecting high-quality training data. Local-perspective meta-evaluation is more suitable for scenarios requiring a fine-grained comparative evaluation, such as labeling data pairs for preference optimization.

\begin{table}[t]
\centering
\small
\setlength{\tabcolsep}{3.4pt}
\renewcommand{\arraystretch}{1.2}
\begin{tabular}{l|cc|cc}
\toprule
\textbf{LLM-as-Judges} & \textbf{$\mathrm{CEM}$} & \textbf{$Acc$} & \textbf{$r$} & \textbf{$\rho$} \\ 
\midrule
GPT-4o  & 0.580(3) & 0.835(4) & 0.562(4) & 0.522(2) \\
GPT-4o mini & 0.393(7) & 0.815(6) & 0.521(7) & 0.507(4) \\
GPT-4 Turbo & 0.741(1) & 0.839(3) & 0.625(1) & 0.513(3) \\
Llama-3.1-70B & 0.402(6) & 0.807(7) & 0.523(6) & 0.491(5) \\
Qwen-2.5-72B & 0.518(5) & 0.840(2) & 0.577(2) & 0.490(7) \\
Gemma-2-27B & 0.527(4) & 0.842(1) & 0.533(5) & 0.490(6) \\
Phi-4-14B & 0.680(2) & 0.815(5) & 0.564(3) & 0.532(1) \\
\bottomrule
\end{tabular}
\caption{The results and rankings of common LLMs as evaluators from our proposed two meta-evaluation perspectives (columns 2 and 3), or using the traditional Pearson ($r$) and Spearman ($\rho$) correlation coefficients (columns 4 and 5).}
\label{tab:preliminary_result}
\end{table}

\begin{table*}[t]
\centering
\small
\renewcommand{\arraystretch}{1.1}
\begin{tabular}{p{2.8cm}p{12cm}}
    \toprule
    \textbf{Evaluation Aspect} & \textbf{Decomposed Sub-aspects} \\ 
    \midrule
    \multirow{4}{=}{\textbf{Coherence}: Measure the quality of all sentences of the summary collectively, to fit together and sound naturally. Consider the quality of the summary as a whole.}
    & \textbf{Logical Flow}: The sentences in the summary are organized in a logical sequence, ensuring smooth and clear transitions between points of the summary in the given order. \\ 
    \cmidrule(l){2-2} 
    & \textbf{Thematic Consistency}: The sentences in the summary revolve around a unified central theme or topic, without unrelated or abrupt information that disrupts continuity. \\ 
    \cmidrule(l){2-2} & \textbf{Referential Clarity}: The references (e.g., pronouns and anaphora) used in the summary should be clear and unambiguous, without incorrect references or cases that the referent does not appear before being referred to. \\ 
    \cmidrule(l){2-2} & \textbf{Sentence Connectivity}: The presence of explicit or implicit connections (e.g., conjunctions, adverbials) between sentences in the summary should be proper and unconfusing. \\ 
    \bottomrule
\end{tabular}
\caption{An example of the evaluation aspect coherence and its decomposed fine-grained sub-aspects in SummEval. The complete information is presented in Tables \ref{tab:subaspect_summeval} and \ref{tab:subaspect_topical} in the appendix.}
\label{tab:subaspect_coherence}
\end{table*}

Although our proposed meta-evaluation approach can be directly applied to existing NLG benchmarks as described above, it has certain limitations in the implementation. In global-perspective meta-evaluation, consistency-based filtering may significantly reduce data volume and lead to an imbalanced rating distribution. In local-perspective meta-evaluation, the target sequence can only be constructed based on the human evaluation scale, lacking sufficient targets to meet the original intention. Moreover, many common NLG evaluation benchmarks face challenges, such as outdated systems and the risk of data contamination. Therefore, we introduce a method of automatically constructing corresponding benchmarks to better implement our dual-perspective meta-evaluation with no need for new human annotations.

\section{Automatic Benchmark Construction}

To address the concerns mentioned in Section \ref{sec:3.3} and construct the required benchmarks for two meta-evaluation perspectives, we adopt the method of controlled error injection on high-quality references. Previous studies \citep{hu-etal-2024-llm, wang2024dhp} have employed perturbation attacks to assess metrics by verifying whether the variations in evaluation results pre- and post-perturbations meet expectations. However, they merely consider whether perturbations exist. In contrast, we further leverage perturbations of varying degrees and quantities while flexibly controlling the error injection to meet different construction requirements.

\subsection{Evaluation Aspect Decomposition}
\label{sec:4.1}

In common NLG benchmarks, the evaluation is often further specified across different evaluation aspects like coherence. We first decompose them into more fine-grained sub-aspects for better guidance of the error injection. Previous research on active evaluation \citep{liu2024hd, li2025dna} has also shown the aspect decomposition can improve the evaluation performance of LLMs, which may play a role similar to the chain of thought \citep{wei2022chain}. To achieve the sub-aspects efficiently, we first prompt OpenAI o1 \citep{openai2024o1preview}, leveraging its strong reasoning ability to generate candidate sub-aspects. Then, they undergo manual selection and refinement, yielding a set of representative sub-aspects, as exemplified in Table \ref{tab:subaspect_coherence}.

\begin{table*}[t]
\centering
\small
\setlength{\tabcolsep}{4.5pt}
\renewcommand{\arraystretch}{1.2}
\begin{tabular}{lcccccccccccc}
\toprule
\multirow{2}{*}{\raisebox{-3.2pt}{\textbf{LLM}}} & \multicolumn{5}{c}{\textbf{SummEval Global}} & \multicolumn{6}{c}{\textbf{Topical-Chat Global}} & \multirow{2}{*}{\raisebox{-3.2pt}{\textbf{Overall}}} \\

\cmidrule{2-6} \cmidrule(lr){7-12} 
 & Coh & Con & Flu & Rel & Avg & Und & Nat & MCtx & Int & UK & Avg \\
\midrule
GPT-4o &  0.765 & 0.708 & 0.702 & 0.801 & 0.744 & 0.978 & 0.815 & 0.875 & 0.792 & 0.888 & 0.870 & 0.807 (5) \\
GPT-4o mini & 0.738 & 0.726 & 0.712 & 0.835 & 0.753 & 0.985 & 0.833 & 0.826 & 0.767 & 0.877 & 0.858 & 0.805 (6) \\
GPT-4 Turbo & 0.731 & 0.691 & 0.690 & 0.785 & 0.724 & 0.972 & 0.774 & 0.841 & 0.849 & 0.889 & 0.865 & 0.795 (7) \\
GPT-3.5 Turbo & 0.673 & 0.667 & 0.653 & 0.779 & 0.693 & 0.905 & 0.772 & 0.699 & 0.829 & 0.844 & 0.810 & 0.751 (12) \\
DeepSeek-V3 & 0.739 & 0.665 & 0.710 & 0.763 & 0.719 & 0.947 & 0.769 & 0.819 & 0.771 & 0.894 & 0.840 & 0.779 (10) \\
Llama-3.1-70B & 0.700 & 0.779 & 0.712 & 0.693 & 0.721 & 0.914 & 0.800 & 0.789 & 0.766 & 0.810 & 0.816 & 0.768 (11) \\
Qwen-2.5-72B & 0.814 & 0.868 & 0.838 & 0.800 & 0.830 & 0.978 & 0.910 & 0.862 & 0.886 & 0.904 & 0.908 & 0.869 (1) \\
Gemma-2-27B & 0.738 & 0.787 & 0.698 & 0.790 & 0.753 & 0.966 & 0.724 & 0.750 & 0.751 & 0.842 & 0.806 & 0.780 (9) \\
Phi-4-14B & 0.761 & 0.721 & 0.775 & 0.752 & 0.752 & 0.963 & 0.823 & 0.837 & 0.887 & 0.901 & 0.882 & 0.817 (4) \\
Auto-J-13B & 0.615 & 0.610 & 0.569 & 0.644 & 0.610 & 0.614 & 0.602 & 0.589 & 0.666 & 0.612 & 0.617 & 0.613 (16) \\
CRITIQUELLM-6B & 0.710 & 0.737 & 0.615 & 0.765 & 0.707 & 0.769 & 0.577 & 0.602 & 0.670 & 0.776 & 0.679 & 0.693 (14) \\
Prometheus-13B & 0.576 & 0.653 & 0.645 & 0.613 & 0.622 & 0.692 & 0.569 & 0.557 & 0.638 & 0.736 & 0.639 & 0.630 (15) \\
Prometheus-2-7B & 0.739 & 0.733 & 0.639 & 0.746 & 0.714 & 0.836 & 0.689 & 0.774 & 0.751 & 0.839 & 0.778 & 0.746 (13) \\
Prometheus-2-8x7B & 0.784 & 0.828 & 0.714 & 0.724 & 0.762 & 0.849 & 0.740 & 0.793 & 0.878 & 0.827 & 0.818 & 0.790 (8) \\
Themis-8B & 0.852 & 0.873 & 0.809 & 0.847 & 0.845 & 0.941 & 0.859 & 0.805 & 0.843 & 0.729 & 0.835 & 0.840 (3) \\
CompassJudger-32B & 0.844 & 0.899 & 0.797 & 0.881 & 0.855 & 0.946 & 0.818 & 0.819 & 0.928 & 0.835 & 0.869 & 0.862 (2) \\
\bottomrule
\end{tabular}
\caption{The results of the $\mathrm{CEM}$ metric for different LLMs on our new global-perspective benchmarks, with the overall performance ranking shown in parentheses. The abbreviations in columns 2-12 represent different evaluation aspects as shown in Tables \ref{tab:subaspect_summeval} and \ref{tab:subaspect_topical}, and the average results of each benchmark.}
\label{tab:global}
\end{table*}

\subsection{Global Perspective}

Since human evaluation is generally regarded as the gold standard, we adhere to the original human evaluation scale to define the different coarse-grained ratings and construct corresponding new targets. Firstly, we generate candidate targets with various qualities by simultaneously injecting different numbers of errors into references using OpenAI o1 and the prompt in Table \ref{tab:ei_prompt_1}. Each error is associated with a random evaluation sub-aspect. Additionally, to ensure both the high quality and diversity of the references, they are generated by GPT-4o with multiple samplings using the prompt in Table \ref{tab:ref_prompt}. We apply a heuristic algorithm to select the most diverse portions based on the similarity measure.

Then, the key lies in the estimation of ratings for these targets. Inspired by the concept of the anchor data used in \citet{liu2024aligning}, we propose an anchoring method that leverages certain existing targets as reference points that align the best with human evaluation preferences. Specifically, we first select the targets from the original benchmark that exhibit high consistency across evaluation results from both humans and strong LLMs, serving as the anchor set $A_r$ for each rating $r$. These targets, also named anchors, are potentially more reliable, and multiple anchors together can be considered approximately representing the quality of the corresponding rating.

Since each candidate target $t$ must belong to a certain rating $r$, it should basically have higher quality than anchor targets of rating $r-1$ and lower quality than anchor targets of rating $r+1$, while exhibiting similar quality to anchor targets of rating $r$. Based on this principle, we compare each candidate target with anchor targets of each rating in a pairwise manner to estimate its rating:
\begin{align*}
    \operatorname*{argmax}_{r}  \frac{ -| \sum_{t_i \in A_r} (F(t \succ t_i) - F(t \prec t_i))|}{|A_{r}|} + \\
    \frac{\sum_{t_i \in A_{r-1}} F(t \succ t_i)}{|A_{r-1}|}+ \frac{\sum_{t_i \in A_{r+1}} F(t \prec t_i)}{|A_{r+1}|}
\end{align*}
\noindent where $| A_r |$ represents the number of targets in anchor set $A_r$, and $F$ denotes the direct pairwise comparison implemented by the LLM, returning 1 if the given quality relationship holds and 0 otherwise. Furthermore, to reduce the number of comparisons and the corresponding computational cost, we use a small part of the data to approximate the error number that corresponds to each rating to guide the more efficient target generation. More details of the anchoring are included in Appendix \ref{sec:B1}.

\subsection{Local Perspective}
\label{sec:4.3}

As for the local perspective, we aim to construct a sequence of targets for each source that follows a clear quality ranking without explicit ratings. To achieve this, we modify the previous error injection method to an iterative process, where we start from the reference and insert one error at a time while unchanging other content. Each error-injected target serves as the input for the next iteration, with each error also randomly corresponding to an evaluation sub-aspect. Since cumulative errors objectively ensure a gradual decline in quality, the sequence from the reference to the last target after the error injection exhibits an order of descending quality. Moreover, free from the constraints of the original human evaluation scale, we can customize the number of iterations to make the sequence contain more targets than the number of different ratings in the scale, thereby ensuring more fine-grained quality differences. The iterative injection process is also implemented by OpenAI o1 with elaborate instructions in Table \ref{tab:ei_prompt_2}. Other details of our benchmark construction are provided in Appendix \ref{sec:B1}.

\section{Experiments}

\subsection{Benchmarks}

We experiment on the two most typical NLG evaluation benchmarks, SummEval \citep{fabbri2021summeval} for text summarization and Topical-Chat \citep{mehri2020usr} for dialogue response generation. Their respective new benchmarks from global and local perspectives are constructed considering their human evaluation scales and task requirements, with their statistics presented in Appendix \ref{sec:B2}. Moreover, we analyze the quality and construction cost of our automatic benchmarks in Appendices \ref{sec:B3} and \ref{sec:B4}, respectively.

\begin{table*}[t]
\centering
\small
\setlength{\tabcolsep}{4.5pt}
\renewcommand{\arraystretch}{1.2}
\begin{tabular}{lcccccccccccc}
\toprule
\multirow{2}{*}{\raisebox{-3.2pt}{\textbf{LLM}}} & \multicolumn{5}{c}{\textbf{SummEval Local}} & \multicolumn{6}{c}{\textbf{Topical-Chat Local}} & \multirow{2}{*}{\raisebox{-3.2pt}{\textbf{Overall}}} \\

\cmidrule{2-6} \cmidrule(lr){7-12} 
 & Coh & Con & Flu & Rel & Avg & Und & Nat & MCtx & Int & UK & Avg \\
\midrule
GPT-4o & 0.661 & 0.707 & 0.661 & 0.646 & 0.669 & 0.776 & 0.714 & 0.807 & 0.702 & 0.595 & 0.719 & 0.694 (2) \\
GPT-4o mini & 0.654 & 0.640 & 0.654 & 0.617 & 0.641 & 0.831 & 0.717 & 0.783 & 0.683 & 0.621 & 0.727 & 0.684 (4) \\
GPT-4 Turbo & 0.683 & 0.723 & 0.643 & 0.643 & 0.673 & 0.764 & 0.721 & 0.776 & 0.640 & 0.624 & 0.705 & 0.689 (3) \\
GPT-3.5 Turbo & 0.609 & 0.615 & 0.569 & 0.601 & 0.598 & 0.700 & 0.617 & 0.698 & 0.690 & 0.555 & 0.652 & 0.625 (10) \\
DeepSeek-V3 & 0.687 & 0.663 & 0.669 & 0.630 & 0.662 & 0.750 & 0.731 & 0.814 & 0.681 & 0.662 & 0.728 & 0.695 (1)\\
Llama-3.1-70B & 0.642 & 0.547 & 0.625 & 0.595 & 0.602 & 0.745 & 0.695 & 0.757 & 0.719 & 0.571 & 0.698 & 0.650 (9) \\
Qwen-2.5-72B & 0.683 & 0.703 & 0.610 & 0.630 & 0.657 & 0.781 & 0.610 & 0.821 & 0.633 & 0.612 & 0.691 & 0.674 (5) \\
Gemma-2-27B & 0.678 & 0.668 & 0.650 & 0.622 & 0.655 & 0.781 & 0.745 & 0.660 & 0.705 & 0.531 & 0.684 & 0.669 (6) \\
Phi-4-14B & 0.636 & 0.695 & 0.659 & 0.617 & 0.652 & 0.774 & 0.624 & 0.726 & 0.643 & 0.550 & 0.663 & 0.658 (8) \\
Auto-J-13B & 0.565 & 0.574 & 0.496 & 0.541 & 0.544 & 0.533 & 0.445 & 0.533 & 0.462 & 0.448 & 0.484 & 0.514 (14) \\
CRITIQUELLM-6B & 0.648 & 0.642 & 0.565 & 0.579 & 0.609 & 0.664 & 0.586 & 0.636 & 0.471 & 0.564 & 0.584 & 0.596 (12) \\
Prometheus-13B & 0.476 & 0.490 & 0.522 & 0.490 & 0.495 & 0.433 & 0.540 & 0.343 & 0.557 & 0.490 & 0.473 & 0.484 (15) \\
Prometheus-2-7B & 0.597 & 0.531 & 0.517 & 0.539 & 0.546 & 0.688 & 0.600 & 0.669 & 0.621 & 0.593 & 0.634 & 0.590 (13) \\
Prometheus-2-8x7B & 0.623 & 0.621 & 0.505 & 0.551 & 0.575 & 0.695 & 0.612 & 0.633 & 0.602 & 0.552 & 0.619 & 0.597 (11) \\
Themis-8B & 0.368 & 0.385 & 0.307 & 0.361 & 0.355 & 0.588 & 0.462 & 0.543 & 0.443 & 0.357 & 0.479 & 0.417 (16) \\
CompassJudger-32B & 0.665 & 0.654 & 0.593 & 0.626 & 0.634 & 0.795 & 0.679 & 0.783 & 0.652 & 0.586 & 0.699 & 0.667 (7) \\
\bottomrule
\end{tabular}
\caption{The results of adjacent pairwise accuracy for different LLMs on new local-perspective benchmarks, with the overall ranking shown in parentheses. The abbreviations represent different evaluation aspects and their averages.}
\label{tab:local}
\end{table*}

\subsection{LLMs for LLM-as-a-Judge}

We select a wide range of representative general-purpose LLMs, as well as those fine-tuned specifically for evaluation, for the LLM-as-a-judge method in our experiments. The general-purpose LLMs include GPT-4o, GPT-4o mini, GPT-4 Turbo, GPT-3.5 Turbo \citep{achiam2023gpt}, DeepSeek-V3 \citep{liu2024deepseek}, Llama-3.1-70B-Instruct \citep{dubey2024llama}, Qwen-2.5-72B-Instruct \citep{qwen2.5}, Gemma-2-27B-Instruct \citep{team2024gemma}, and Phi-4-14B \citep{abdin2024phi}. The specifically fine-tuned LLMs include Auto-J-13B \citep{ligenerative}, CRITIQUELLM-6B \citep{ke2024critiquellm}, Prometheus-13B \citep{kim2023prometheus}, Prometheus-2-7B, Prometheus-2-8x7B \citep{kim2024prometheus}, Themis-8B \citep{hu2024themis} and CompassJudger-32B \citep{cao2024compassjudger}. The detailed experimental settings, including the evaluation instructions, are described in Appendix \ref{sec:A4}.

\begin{figure}[t]
    \centering
    \begin{subfigure}[b]{0.237\textwidth}
        \centering
        \includegraphics[width=\linewidth]{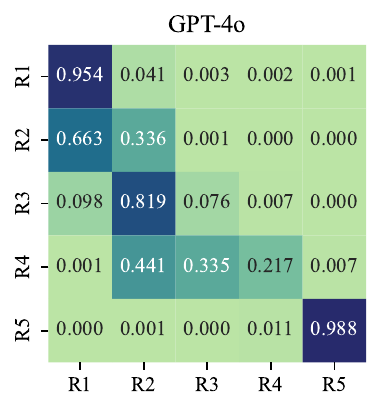}
    \end{subfigure}
    \hfill
    \begin{subfigure}[b]{0.237\textwidth}
        \centering
        \includegraphics[width=\linewidth]{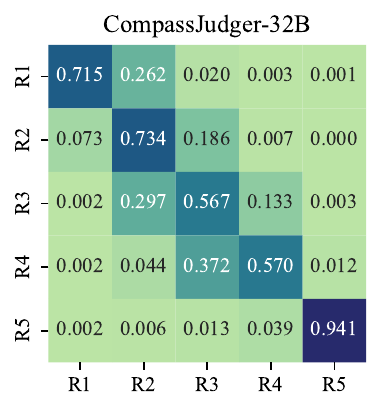}
    \end{subfigure}
    \caption{The confusion matrices of two LLMs on SummEval, where the vertical and horizontal axes represent the true and predicted ratings from 1 to 5, respectively.}
    \label{tab:confusion}
\end{figure}

\subsection{Global Perspective}

We present the performance of different LLMs as the evaluators on our new benchmarks of SummEval and Topical-Chat from the global perspective in Table \ref{tab:global}. The results show that Qwen-2.5-72B and CompassJudger-32B achieve the best overall performance, both greatly surpassing GPT-4o. Furthermore, Phi-4-14B and GPT-4o mini, despite the relatively small number of parameters, exhibit close performance to GPT-4o. This indicates that current small-scale general-purpose LLMs have already achieved a considerable level of coarse-grained evaluation capability, even without specialized fine-tuning for evaluation scenarios.

To further understand the evaluation behaviors of different LLMs, we generate the confusion matrices. Figure \ref{tab:confusion} shows two representative LLMs on SummEval, with more results illustrated in Appendix \ref{sec:C1}. While GPT-4o is more accurate in judging the targets of the best and worst quality, it tends to be overly stringent when evaluating the targets of medium quality, often considering them as the lower ratings. In contrast, CompassJudger-32B, though less accurate for low-quality targets, demonstrates a more balanced performance across different ratings. Such discrepancy may stem from the different expectations regarding text quality among LLMs, thereby affecting their evaluation performance. For example, the superior linguistic capabilities of GPT-4o may lead it to impose overly high standards during evaluation.

\subsection{Local Perspective}

In our local-perspective benchmarks, the targets constructed for each source contain different numbers of cumulative errors as described in Section \ref{sec:4.3}. We define the pair of targets of the same source, with one containing $k$ more errors than the other, as the target pair with $I(k)$. So the performance of different LLMs is assessed on target pairs with $I(1)$ according to Section \ref{sec:3.2}, with the results presented in Table \ref{tab:local}. Compared with the global perspective, the ranking of LLMs has changed significantly. For instance, GPT-4 Turbo and DeepSeek-V3, which perform at a mid-tier level in global-perspective benchmarks, now achieve the top overall performance, and GPT-4o overtakes both Qwen-2.5-72B and CompassJudger-32B. Therefore, there is a clear difference between the LLMs' proficiency in these two evaluation capabilities, highlighting the necessity of distinguishing them and the effectiveness of our proposed meta-evaluation framework.

\begin{figure}[t]
    \centering
    \includegraphics[width=\linewidth]{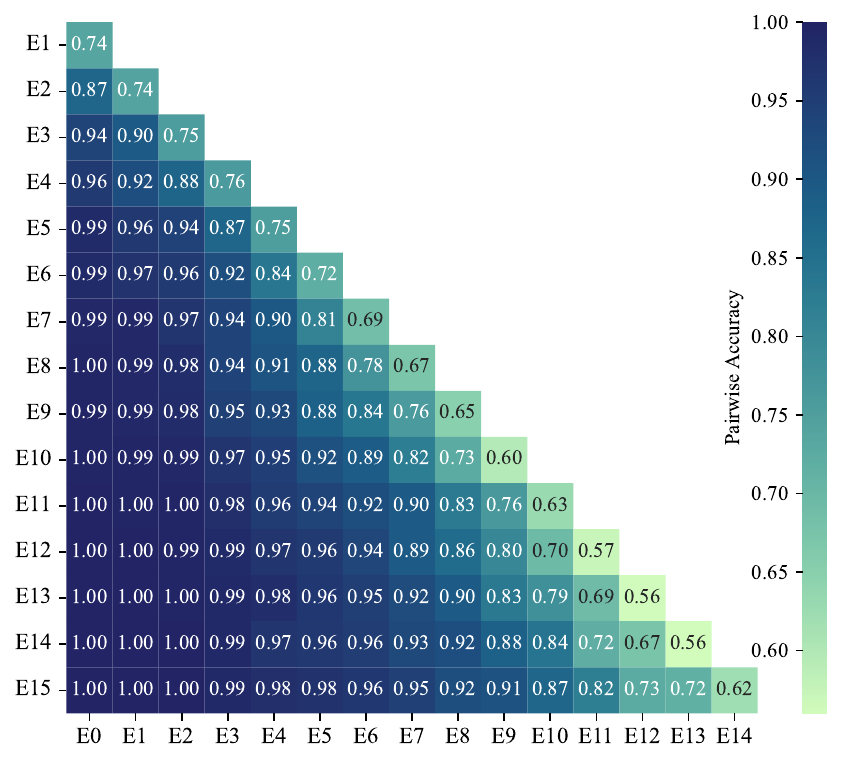}
    \caption{The complete results of all target pairs, where the diagonal corresponds to target pairs with $I(1)$.}
    \label{fig:errorheat}
\end{figure}

\begin{table}[t]
\centering
\small
\setlength{\tabcolsep}{8pt}
\renewcommand{\arraystretch}{1.2}
\begin{tabular}{lcc}
\toprule
\textbf{LLM} & \textbf{Scoring} & \textbf{Comparison} \\
\midrule
GPT-4o & 0.669 & 0.401\\
GPT-4o mini & 0.641 & 0.500 \\
GPT-4 Turbo & 0.673 & 0.533 \\
GPT-3.5 Turbo & 0.599 & 0.341 \\
DeepSeek-V3 & 0.662 & 0.654 \\
Llama-3.1-70B & 0.602 & 0.499 \\
Qwen-2.5-72B & 0.657 & 0.553 \\
Gemma-2-27B & 0.655 & 0.388 \\
Phi-4-14B & 0.652 & 0.561 \\
Auto-J-13B & 0.544 & 0.563 \\
CRITIQUELLM-6B & 0.609 & 0.398 \\
Prometheus-2-7B & 0.546 & 0.567 \\
Prometheus-2-8x7B & 0.575 & 0.694\\
CompassJudger-32B & 0.634 & 0.639\\
\bottomrule
\end{tabular}
\caption{The results of different LLMs using two evaluation approaches on the local-perspective benchmark of SummEval: scoring two targets separately followed by comparing the scores, and direct pairwise comparison.}
\label{tab:comparison}
\end{table}

Moreover, we present the complete pairwise accuracy on all target pairs in Figure \ref{fig:errorheat}, covering various quality combinations. We take GPT-4o on SummEval as an example, while other cases follow a similar pattern, as shown in Appendix \ref{sec:C1}. The horizontal and vertical axes represent the cumulative error counts of two targets in the pair, which reflect their respective qualities. Therefore, in Figure \ref{fig:errorheat}, target pairs located further toward the upper left have higher qualities, while those further toward the lower left exhibit greater quality differences. The results indicate that LLMs have better discriminative ability on target pairs with higher qualities and greater quality differences.

\begin{figure*}[t]
    \centering
    \begin{subfigure}[b]{0.329\textwidth}
        \centering
        \includegraphics[width=\linewidth]{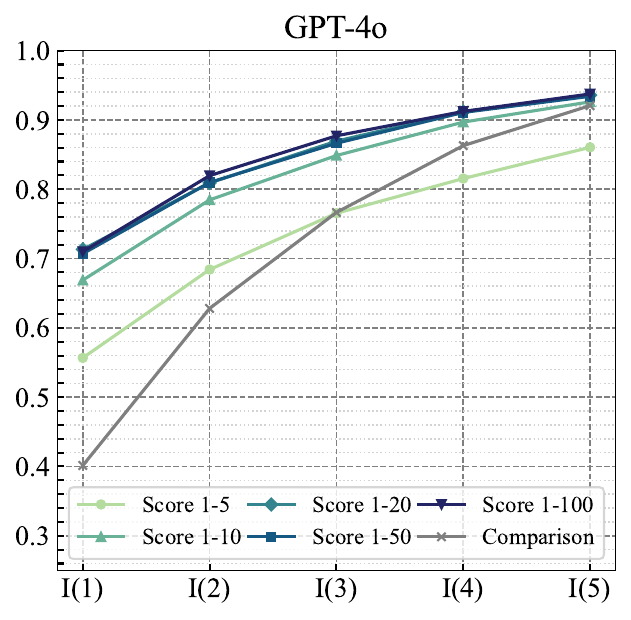}
    \end{subfigure}
    \hfill
    \begin{subfigure}[b]{0.329\textwidth}
        \centering
        \includegraphics[width=\linewidth]{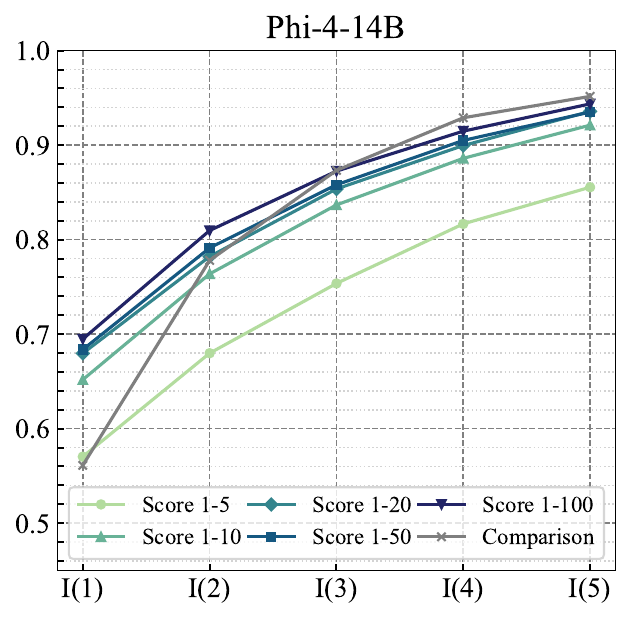}
    \end{subfigure}
    \hfill
    \begin{subfigure}[b]{0.329\textwidth}
        \centering
        \includegraphics[width=\linewidth]{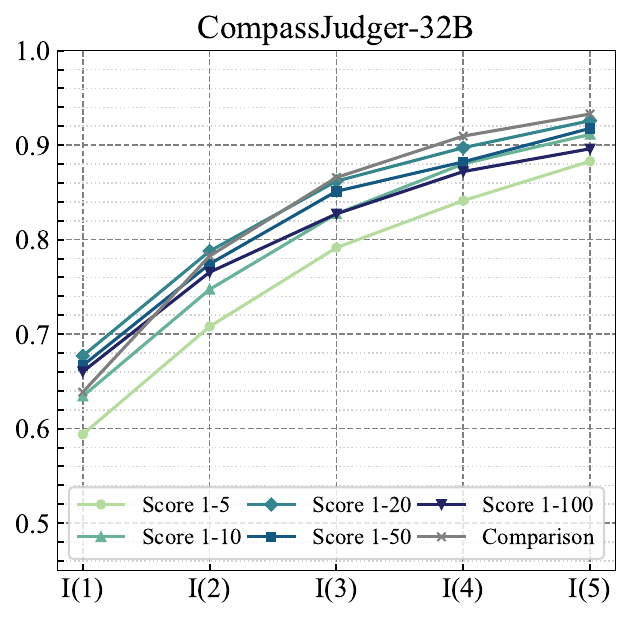}
    \end{subfigure}
    \caption{Evaluation results of LLMs using approaches of directly comparing and scoring with different ranges on target pairs with five levels of differences, which increase from $I(1)$ to $I(5)$.}
    \label{tab:linechart}
\end{figure*}

Furthermore, we experiment with a more intuitive evaluation approach, where LLMs directly compare each target pair with $I(1)$. As shown in Table \ref{tab:comparison}, it is surprising that the LLMs except those specifically fine-tuned perform worse with direct pairwise comparison than the previous evaluating each target separately and then comparing the scores, which contradicts observations from prior studies \citep{liusie2024llm, chenmllm}. We suppose this is because different evaluation approaches have their respective limits on discriminative power, and differences within the target pairs with $I(1)$ are subtle. Therefore, we further explore target pairs with greater differences up to $I(5)$, as well as other scoring ranges (e.g., 1–100) to observe their influence on the evaluation performance. The results in Figure \ref{tab:linechart} show that the direct comparison approach is only competitive to the scoring approach when differences between targets are significant. Moreover, LLMs exhibit similar performance when the scoring range extends beyond 1–10, and the optimal range varies.

\section{Related Work}

As LLM-as-a-judge methods become increasingly prevalent, prior research has introduced various meta-evaluation benchmarks to assess their effectiveness. Most of these benchmarks emphasize the correlation between LLM-generated evaluations and human judgments \citep{DBLP:conf/iclr/YeKKHKJTKS24,DBLP:journals/corr/abs-2406-05761,DBLP:conf/nips/ZhengC00WZL0LXZ23,DBLP:conf/iclr/ZengYG0G024,DBLP:journals/corr/abs-2308-01862}. Taking a slightly different approach, \citet{DBLP:journals/corr/abs-2406-12624} evaluate LLMs using an objective, knowledge-based question-answering task. \citet{DBLP:journals/corr/abs-2410-12784} further propose a pipeline for automatically constructing benchmarks that focus solely on factual and logical correctness.

In the context of NLG tasks, \citet{DBLP:journals/corr/abs-2406-18403} explore the performance of various LLM-based evaluators across 20 NLP tasks, including translation and summarization. Other studies have targeted specific NLG tasks, such as translation \citep{DBLP:conf/wmt/FreitagMDLAR0BK24}, summarization \citep{DBLP:conf/naacl/LiuFCZHJLRWC24,DBLP:conf/acl/SiledarNMRNBBPS24}, dialogue generation \citep{DBLP:conf/emnlp/ZhangDTST023,DBLP:conf/emnlp/MendoncaTL24}, and story generation \citep{DBLP:journals/tacl/ChhunSC24}.

Our work differs from these studies in a fundamental way: we decompose the widely used—yet often ambiguous—measure of human-LLM correlation into two clearer components: the global perspective and local perspective, each with its own dedicated evaluation metric. Additionally, we propose an automatic data construction method to instantiate our meta-evaluation framework. Our approach exposes the limitations of relying solely on human-LLM correlation in NLG meta-evaluation and broader LLM-as-a-judge applications, and introduces a more interpretable alternative.

Beyond human-LLM correlation, \citet{DBLP:conf/acl/WangLCCZLCKLLS24} investigate position bias in pairwise LLM comparisons. \citet{DBLP:journals/corr/abs-2410-02736} employ perturbation methods to quantify 12 distinct biases in LLM-based evaluators. \citet{DBLP:conf/coling/LeeHT25} examine the consistency of LLM evaluations under multiple sampling strategies and varying scoring scales. \citet{zhao-etal-2024-measuring} explore additional desiderata for pairwise evaluation, such as transitivity. \citet{DBLP:journals/corr/abs-2408-13704} design perturbation tests across six NLG tasks to benchmark LLM evaluation capabilities. In contrast to these works, our study remains focused on the correlation between LLMs and human judgments in evaluation tasks.

More closely related to our work are studies in machine translation that seek to redefine the notion of correlation between automatic evaluation metrics and human judgments. However, such efforts are relatively rare and differ significantly in motivation from ours. Specifically, \citet{DBLP:conf/emnlp/Perrella0CBN24} highlight that in machine translation (MT) meta-evaluation, correlation lacks interpretability in terms of range consistency, error attribution, and performance. To address this, they reformulate meta-evaluation as a binary classification and re-ranking task. In contrast, \citet{DBLP:conf/emnlp/AgrawalFRM24} redefine MT meta-evaluation as assessing whether automatic evaluation metrics can effectively distinguish high-quality translations.

\section{Conclusion}

In this work, we propose a new dual-perspective NLG meta-evaluation framework to address certain limitations in the traditional paradigm. Our framework focuses on different evaluation capabilities from global and local perspectives, thereby offering better interpretability. Moreover, we introduce a method of automatically constructing corresponding new benchmarks, avoiding potential data contamination and the cost of new human annotations. We further conduct comprehensive experiments with 16 representative LLMs as evaluators, analyzing their proficiency in different evaluation capabilities. Our proposed framework and study findings aim to provide new insights for NLG meta-evaluation and promote further research.

\section*{Limitations}

In constructing our benchmarks, we leverage some LLMs for certain steps, such as generating candidate targets. However, we do not rely entirely on more powerful OpenAI o1, meaning the construction may not be optimal. This decision is driven by the high cost of OpenAI o1 and that some tasks can be satisfactorily handled by more affordable LLMs, such as GPT-4o. Nonetheless, our construction method inherently benefits from advancements in LLMs. As their capabilities improve and inference costs decline, the quality and cost-effectiveness of our benchmark will naturally improve.


\section*{Ethics Statement}

This work does not pose any ethical issues. All datasets, open-source LLMs, and API calls used in our work are publicly available. And we comply with their respective licenses and use them only for research purposes. The error injection methods employed in our benchmark construction focus exclusively on text quality and do not introduce any harmful content or personally sensitive information. Additionally, we responsibly recruit annotators who have a good command of English from local universities for our human evaluation and provide reasonable payment for their work.

\section*{Acknowledgements}
This work was supported by Beijing Science and Technology Program (Z231100007423011) and Key Laboratory of Science, Technology and Standard in Press Industry (Key Laboratory of Intelligent Press Media Technology). We appreciate the anonymous reviewers for their helpful comments. Xiaojun Wan is the corresponding author.

\bibliography{custom}

\appendix

\section{Experimental Details}

\subsection{Human Re-Evaluation in Section 3}
\label{sec:A1}

For the first test group, we randomly sample the target pairs from the benchmark that are rated $(r, r, r)$ and $(r, r, r \pm 1)$ by three human annotators, representing the quality of an original rating $r$ and a closest new rating of it. For the second test group, we randomly sample the target pairs from the benchmark that are rated $(r, r, r)$ and $(r + 1, r + 1, r + 1)$ by three human annotators, representing the quality of two closest original ratings. We sample 50 targets for each group as per the above requirements, with the order of each pair being randomly shuffled. Then, we recruit two human evaluators familiar with NLG evaluation to independently judge which target in each pair has higher quality. These judgments are finally compared with the existing averaged ratings from the benchmark, with the results from two annotators being averaged.

\subsection{Ordinal Classification Metric}
\label{sec:A2}

Given an ordinal classification task that contains $C$ ordinal classes represented by consecutive integers, let \(c_{ij}\) denote the number of items whose gold class is \( j \), predicted as \( i \) (\( i, j \in C \)), and \(n_i\) denote the number of items whose gold class is \( i \), and \(N\) is the total number of items. Then \textit{Closeness Evaluation Measure} ($\mathrm{CEM}$) is defined as:
\begin{equation*}
    \mathrm{CEM} = \frac{\sum_{i\in C}\sum_{j\in C}c_{ij}\mathrm{prox}(i,j)}{\sum_{i\in C}n_i\mathrm{prox}(i,i)}
\end{equation*}
where $\mathrm{prox}(i, j) = -\log \left (\frac{\sum_{k=a}^bn_k - \frac{n_i}{2}}{N} \right)$, with \(a=\min(i,j)\) and \(b=\max(i,j)\), reflects the \textit{informational closeness} that the metric assigns to a pair of classes \(i, j\). \citet{amigo2020effectiveness} conduct extensive analysis and find that $\mathrm{CEM}$ demonstrates good evaluation capability and robustness on ordinal classification, surpassing other common metrics like accuracy and correlation.

\begin{table}[t]
\centering
\small
\setlength{\tabcolsep}{3pt}
\renewcommand{\arraystretch}{1.1}
\begin{tabular}{ccccc}
\toprule
& \textbf{Coherence} & \textbf{Consistency} & \textbf{Fluency} & \textbf{Relevance} \\
\midrule
Rating 1 & 13 & 14 & 5 & 5 \\
Rating 2 & 29&14&10&19 \\ 
Rating 3&33&5&24&37 \\
Rating 4&14&1&2&84 \\
Rating 5&111&1306&1150&83 \\
\bottomrule
\end{tabular}
\caption{The rating distribution of the retained data on SummEval after the consistency-based data filtering.}
\label{tab:distribution}
\end{table}

\subsection{Preliminary Study on SummEval}
\label{sec:A3}

For the traditional NLG meta-evaluation paradigm, we use the entire SummEval dataset to compute the dataset-level Pearson and Spearman correlations between the evaluation results of LLMs and humans, like $\rho (\{h_{ij}\}_{i=1,j=1}^{n,m}, \{x_{ij}\}_{i=1,j=1}^{n,m})$. The evaluation scores from LLMs are obtained using a prompt similar to Table \ref{tab:SC_prompt} with the scoring range of 1-5, and we average the results from ten samplings with a temperature of 1.

For our proposed new meta-evaluation framework, we retain only the targets where all three human annotators provide consistent ratings in the original benchmark. For global-perspective meta-evaluation, the results are calculated on an ordinal classification task with five categories and $\mathrm{CEM}$ metric. For local-perspective meta-evaluation, since we lack a fine-grained measure for quality, the target sequence is simply constructed from targets with different original ratings, which actually contradicts our original intent. The corresponding experimental settings are similar to those described in Appendix \ref{sec:A4}.

Moreover, after consistency-based data filtering, the retained data for coherence and relevance is reduced to less than 20\% of the original benchmark (1600 for each aspect), while consistency and fluency retain relatively sufficient data. However, their rating distributions are highly imbalanced, as shown in Table \ref{tab:distribution}.

\subsection{Testing LLMs on New Benchmarks}
\label{sec:A4}

In the global-perspective meta-evaluation, we use the prompts in Table \ref{tab:OC_prompt} to generate evaluation ratings for general-purpose LLMs, keeping the evaluation scale consistent with the corresponding benchmark. For specifically fine-tuned LLMs, since they have been trained with their specific instructions, they do not follow the prompts in Table \ref{tab:OC_prompt}. So they are prompted with their original evaluation instructions and scales. All the LLMs generate ten results at a temperature of 1 with multiple samplings, and the mode of these results is taken as the final evaluation rating. In particular, the ratings of some specifically fine-tuned LLMs are uniformly rescaled to match the scale of the corresponding benchmark.

In the local-perspective meta-evaluation, we use the prompts in Table \ref{tab:SC_prompt} for general-purpose LLMs, while the specifically fine-tuned LLMs are prompted with their specific instructions. To obtain high-precision evaluation scores, we follow \citet{chiang-lee-2023-closer} to average ten evaluation results from multiple samplings. The scoring range is set to 1-10 in our main experiments, which is sufficient according to our tests. Additionally, for the direct pairwise comparison, we apply the prompts in Table \ref{tab:PC_prompt} for general-purpose LLMs. For the specifically fine-tuned LLMs that can conduct the pairwise comparison, their specific instructions are still employed. To mitigate the well-known position bias, we adopt the common approach of swapping the order of target pairs and aggregating the judgments from both orderings and ten samplings.

\section{An Example for Limitations Discussed in Section 3}
\label{sec:example}

\begin{table}[t]
    \centering
    \small
    \setlength{\tabcolsep}{5pt}
    \begin{tabular}{l|c|c|c|c}
        \toprule
         & Target A & Target B & Target C & Target D \\ 
        \midrule
        Evaluator & \multirow{2}{*}{(2, 3, 4)} & \multirow{2}{*}{(3, 3, 3)} & \multirow{2}{*}{(3, 3, 3)} & \multirow{2}{*}{(3, 3, 4)} \\ 
        Scale 1-5 &  &  &  &  \\ 
        \midrule
        Expert & \multirow{2}{*}{(6, 6, 6)} & \multirow{2}{*}{(5, 5, 5)} & \multirow{2}{*}{(6, 6, 6)} & \multirow{2}{*}{(6, 6, 6)} \\ 
        Scale 1-10 &  &  &  &  \\ 
        \midrule
        Metric & 6.1 & 4.9 & 6.2 & 5.9 \\ 
        \bottomrule
    \end{tabular}
    \caption{An example of four targets with different cases of ratings and qualities.}
    \label{tab:example1}
\end{table}

We assume there are four targets in an evaluation benchmark, each rated on a 1-5 scale by three crowdsourced evaluators as shown in Table \ref{tab:example1}. Additionally, we further assume ideally that there are three expert evaluators rating on a more fine-grained 1-10 scale and their evaluations are accurate. Then Target A and Target B correspond to the issue of \textit{"The targets with the same averaged rating do not necessarily have the same quality"}, while Target C and Target D correspond to the issue of \textit{"The additional quality levels generated by averaging aggregation are not necessarily valid."} Furthermore, although Target B and Target C have the same ratings from crowdsourced evaluators, they can still be distinguished when assessed on a more fine-grained scale. Since the quality of targets is continuously distributed, such a situation is inevitable. In this case,  although the metric in Table \ref{tab:example1} correctly judges that Target C is better than Target B with its evaluation scores, it will be penalized in the calculation of correlation measures.

\section{Details of New Automatic Benchmarks}

\begin{table}[t]
\centering
\small
\setlength{\tabcolsep}{8pt}
\renewcommand{\arraystretch}{1.1}
\begin{tabular}{lc}
\toprule
\textbf{LLM} & \textbf{Accuracy} \\
\midrule
GPT-4o & 0.945 \\
GPT-4o mini & 0.934 \\
GPT-4 Turbo & 0.911\\
Qwen-2.5-72B & 0.860 \\
Phi-4-14B & 0.874 \\
CompassJudger-32B & 0.920\\
\bottomrule
\end{tabular}
\caption{The accuracy of different LLMs on our constructed anchor set on SummEval. The LLMs are required to directly compare each pair of targets that have different ratings.}
\label{tab:anchor_acc}
\end{table}

\begin{table*}[t]
\centering
\small
\renewcommand{\arraystretch}{1.1}
\begin{tabular}{lccccccc}
    \toprule
    \multirow{2}{*}{\centering \textbf{Benchmark}} & \multirow{2}{*}{\centering \textbf{\#Aspect}} &
    \multirow{2}{*}{\centering \textbf{\#Annotator}} &
    \multirow{2}{*}{\centering \textbf{Evaluation Scale}} & 
    \multirow{2}{*}{\centering \textbf{\#Source}} &
    \multicolumn{3}{c}{\textbf{\#Target per Source}} \\
    \cmidrule{6-8}
    & & & & & \textbf{Original} & \textbf{Global} & \textbf{Local} \\
    \midrule
    SummEval & 4 & 3 & 1-5 & 100 & 16 & 15 & 16 \\
    Topical-Chat & 5 & 3 & 0-1 \& 1-3 & 60 & 6 & 10 & 8 \\
    \bottomrule
\end{tabular}
\caption{Detailed data statistics of the original and our new benchmarks of SummEval and Topical-Chat.}
\label{tab:data}
\end{table*}

\subsection{Construction Process}
\label{sec:B1}

Since the references in many existing NLG evaluation benchmarks are rule-based rather than human-written, we use the prompts in Table \ref{tab:ref_prompt} and GPT-4o for generating reference targets for each source to ensure the quality. For global-perspective and local-perspective meta-evaluation, we employ OpenAI o1 to conduct the error injection using the prompts in Tables \ref{tab:ei_prompt_1} and \ref{tab:ei_prompt_2}, respectively. Each error type is randomly selected from the decomposed evaluation sub-aspects in Tables \ref{tab:subaspect_summeval} and \ref{tab:subaspect_topical}.

More specifically, in the global-perspective meta-evaluation, for each source $s_i$ and each coarse-grained rating $r$, we construct $j$ targets ${t_{i1}^r, \cdots, t_{ij}^r,}$ through the rating estimation process described in Section 4.2, ensuring as uniform distribution as possible. The corresponding benchmark consists of data as follows:
\begin{equation*}
    \{s_i, \{t_{i1}^r, , t_{i2}^r, \cdots, t_{ij}^r\}_{r=1}^l\}_{i=1}^n 
\end{equation*}
In the local-perspective meta-evaluation, for each source $s_i$, we construct a sequence of $k$ targets $t_{i1}, t_{i2}, \cdots, t_{ik}$ with increasing cumulative error counts, ensuring a decreasing quality order. The corresponding benchmark consists of data as follows:
\begin{equation*}
    \{s_i, \{t_{i1}, t_{i2}, \cdots, t_{ik}\}\}_{i=1}^n 
\end{equation*}

Based on the results of the preliminary study in Section 3.3 and the general capabilities of LLMs, we select GPT-4o for constructing the anchor set and GPT-4o-mini as the comparator for rating estimation in global meta-evaluation. In particular, the target pairs to be compared here may correspond to different sources, so some LLMs that have been fine-tuned for specific comparative evaluation, such as Prometheus-2-8x7B, are unavailable. Additionally, although we find that many LLMs' performance of the pairwise comparison is only competitive when there are significant differences within target pairs in Section 5.4, the comparison scenarios here involve coarse-grained ratings that meet the requirements. We present the comparison accuracy of different available LLMs on the anchor set built on summEval in Table \ref{tab:anchor_acc}, and GPT-4o mini shows the best cost-effectiveness. 

Moreover, the selection of anchor targets prioritizes those with high consistency among evaluation results from humans and LLMs:
\begin{align*}
    \frac{1}{a}\sum_{k=1}^a|h_{ij}^k - r| + & \frac{1}{b}\sum_{k=1}^b|x_{ij}^k - r| \\
    + & |\frac{1}{a}\sum_{k=1}^ah_{ij}^k - \frac{1}{b}\sum_{k=1}^bx_{ij}^k|
\end{align*}
\noindent For both SummEval and Topical-Chat, we select five anchor targets for each evaluation aspect and each corresponding rating, with $a$ and $b$ equaling 3 and 10, respectively.

\subsection{Benchmark Statistics}
\label{sec:B2}

When constructing the new automatic evaluation benchmark, we consider the characteristics of the original benchmark. In the global-perspective meta-evaluation, the number of targets for each source is similar to that in the original benchmark, and the distribution of their ratings is as balanced as possible. In the local-perspective meta-evaluation, taking into account the human evaluation scales from the original benchmark and the target length of the corresponding NLG task, we set the number of targets contained in the target sequence for each source to approximately three times that of the different original ratings. In addition, we keep the sources of the new benchmark the same as those of the original benchmark, since the primary objects of meta-evaluation are newly constructed targets. In total, there are 6000 and 6400 targets for new global-perspective and local-perspective benchmarks of SummEval, respectively, and 3000 and 2400 targets for new corresponding benchmarks of Topical-Chat, respectively. The detailed statistics are shown in Table \ref{tab:data}.

\subsection{Benchmark Quality}
\label{sec:B3}

In Table \ref{tab:anchor}, we present the overall comparison results between newly constructed targets and the corresponding targets in the anchor set, based on the estimation method introduced in Section 4.2. More specifically, for the new target $t$ with the rating $r$, columns of $t^{r-1}\prec t$ and $t \prec t^{r + 1}$ represent the probability that $t$ has a higher quality than targets $t^{r-1}$ with the rating $r-1$ in the anchor set $A_{r-1}$ and the probability that $t$ has a lower quality than targets $t^{r+1}$ with the rating $r+1$ in the anchor set $A_{r+1}$, respectively. Both these two probabilities are ideally as high as possible. Columns of $t^{r}\prec t$ and $t \prec t^r$, on the other hand, indicate the probability that $t$ has a higher and lower quality than targets $t^r$ with the same rating $r$ in the anchor set $A_r$, respectively, and these two probabilities are expected to be as close as possible. The results demonstrate that our constructed global meta-evaluation benchmarks basically align with our expectations.

\begin{table}[t]
\centering
\small
\setlength{\tabcolsep}{3pt}
\renewcommand{\arraystretch}{1.1}
\begin{tabular}{lcccc}
\toprule
\textbf{Aspect} & $t^{r-1}\prec t$ & $t^{r}\prec t$ & $t \prec t^r$ & $t \prec t^{r + 1}$ \\
\midrule
\multicolumn{5}{c}{\emph{SummEval}} \\
Coherence & 0.905 & 0.482 & 0.437 & 0.937 \\
Consistency & 0.881 & 0.464 & 0.450 & 0.902 \\
Fluency & 0.839 & 0.472 & 0.445 & 0.961 \\
Relevance & 0.970 & 0.390 & 0.544 & 0.944 \\
\midrule
\multicolumn{5}{c}{\emph{Topical-Chat}} \\
Understandability & 1.000 & 0.621 & 0.339 & 0.991 \\
Naturalness & 1.000  & 0.333 & 0.582 & 0.993 \\
Context Maintanence & 0.954 & 0.339 & 0.555 & 0.959 \\
Interestingness & 0.983 & 0.388 & 0.544 & 0.993 \\
Knowledge Use & 1.000 & 0.255 & 0.685 & 1.000 \\ 
\bottomrule
\end{tabular}
\caption{The comparison results between newly constructed targets and the anchor targets during rating estimation in constructing global-perspective benchmarks.}
\label{tab:anchor}
\end{table}

To further validate the quality of our constructed benchmarks, we conduct a human evaluation. We still recruit the two human annotators who previously performed the re-evaluation and sample 100 targets each from the global-perspective and local-perspective meta-evaluation benchmarks of SummEval. The human annotators are required to evaluate these 200 targets independently. For the global-perspective benchmark, they should judge the rating on a scale of 1 to 5, while for the local-perspective benchmark, they perform the pairwise comparison in the pair of adjacent targets. The final human evaluation results indicate an overall accuracy of 84.5\%, demonstrating that our benchmarks are of satisfactory quality.

\subsection{Construction Cost}
\label{sec:B4}

\begin{table}[t]
\centering
\small
\renewcommand{\arraystretch}{1.1}
\setlength{\tabcolsep}{3.5pt}
\begin{tabular}{lccc}
\toprule
\textbf{Step} & \textbf{LLM} & \textbf{\#API Call} & \textbf{Cost} \\ 
\midrule
Reference Generation & GPT-4o & 10K & \$6 \\
Error Injection (Global) & OpenAI o1 & 8K & \$132 \\
Error Injection (Local) & OpenAI o1 & 6K & \$90 \\
Rating Estimation & GPT-4o mini & 1.3M & \$98 \\
\bottomrule
\end{tabular}
\caption{The main construction cost of our new benchmarks for SummEval.}
\label{tab:cost}
\end{table}

The construction cost of our automatic benchmarks is primarily concentrated on API calls for LLMs, covering steps such as reference generation, error injection, and rating estimation. Taking SummEval as an example, Table \ref{tab:cost} shows the detailed cost associated with each step. In the construction of our global-perspective benchmark, we generate more candidate targets than required for each rating, allowing us to select the optimal ones according to our requirements. The total cost of constructing two new benchmarks on SummEval amounts to only about \$326, which is significantly lower than the expected cost of the corresponding human annotation, and the case of Topical-Chat is similar.

\begin{table*}[!htp]
  \centering\small
  \renewcommand{\arraystretch}{1.1}
  \begin{tabular}{p{15cm}}
  \toprule
  \textbf{SummEval} \\
  \midrule
    Write a summary for the given news article in three or four sentences. \newline
    The summary should be well-written and include all and only the important information from the news article, without any unnecessary, fabricated, and incorrect information. \newline
    \newline
    Article: \newline
    \{source\} \newline
    Summary: \\
  \midrule
  \textbf{Topical-Chat} \\
  \midrule
    Generate a next-turn response for a dialogue context between two people. \newline
    The response must be conditioned on the given fact and use the fact well. (e.g., the response mentions or refers to the given fact appropriately.) \newline
    The response should be understandable, naturally written, and on the conversation's topic. \newline
    \newline
    Fact: \newline
    \{addition\} \newline
    Dialogue Context: \newline
    \{source\} \newline
    Response: \\
  \bottomrule
  \end{tabular}
  \caption{Prompts for reference generation in benchmark construction using GPT-4o.}
  \label{tab:ref_prompt}
\end{table*}

\section{Additional Experimental Results}
\label{sec:C1}

\begin{figure*}[ht]
    \centering
    \begin{subfigure}[b]{0.245\textwidth}
        \centering
        \includegraphics[width=\linewidth]{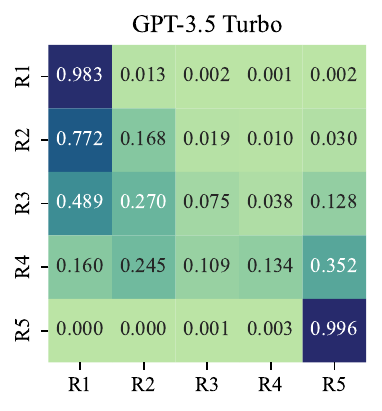}
    \end{subfigure}
    \hfill
    \begin{subfigure}[b]{0.245\textwidth}
        \centering
        \includegraphics[width=\linewidth]{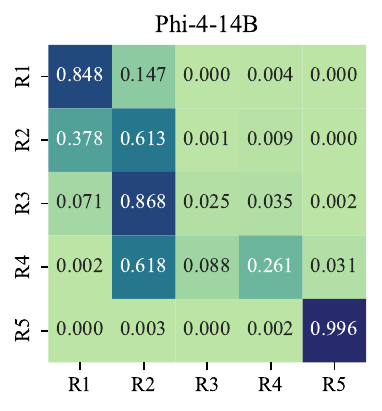}
    \end{subfigure}
    \hfill
    \begin{subfigure}[b]{0.245\textwidth}
        \centering
        \includegraphics[width=\linewidth]{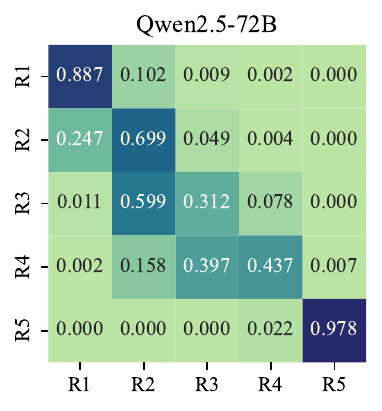}
    \end{subfigure}
    \hfill
    \begin{subfigure}[b]{0.245\textwidth}
        \centering
        \includegraphics[width=\linewidth]{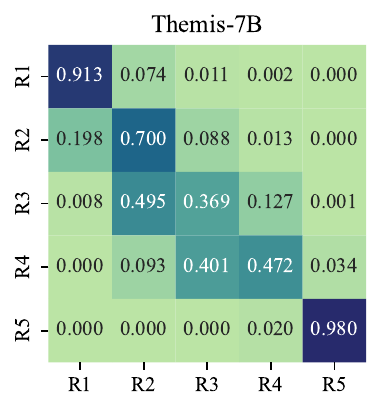}
    \end{subfigure}
    \caption{The confusion matrices of LLMs on our global-perspective benchmark of SummEval, where the vertical and horizontal axes represent the true and predicted ratings from 1 to 5, respectively.}
    \label{tab:confusion_more}
\end{figure*}

\begin{figure*}[t]
    \centering
    \begin{subfigure}[b]{0.245\textwidth}
        \centering
        \includegraphics[width=\linewidth]{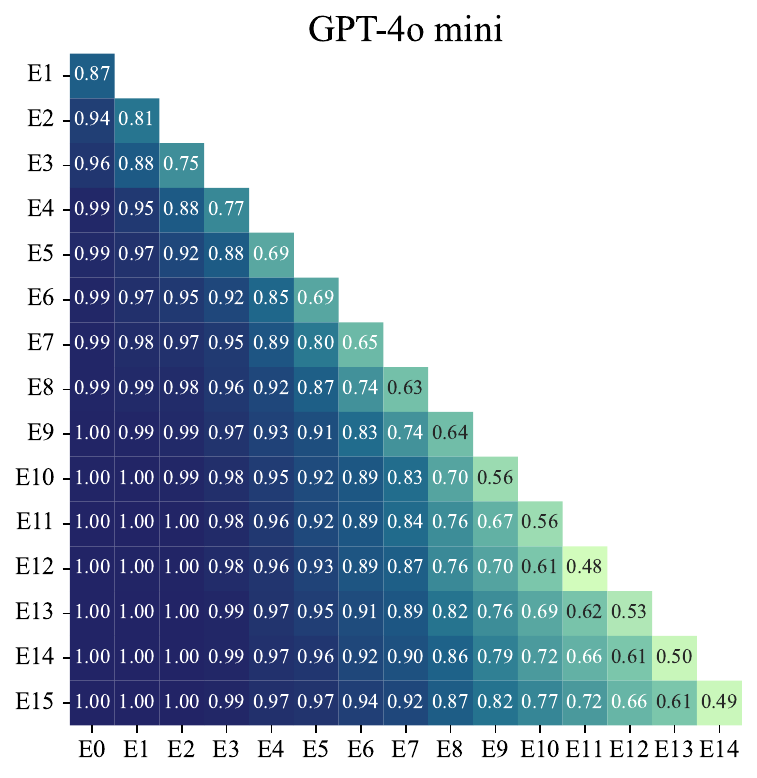}
    \end{subfigure}
    \hfill
    \begin{subfigure}[b]{0.245\textwidth}
        \centering
        \includegraphics[width=\linewidth]{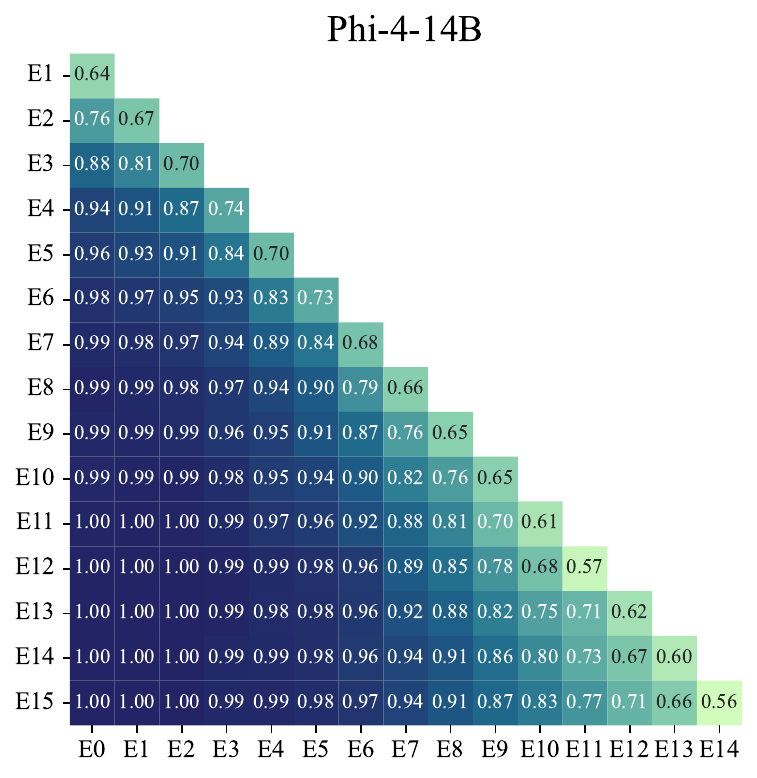}
    \end{subfigure}
    \hfill
    \begin{subfigure}[b]{0.245\textwidth}
        \centering
        \includegraphics[width=\linewidth]{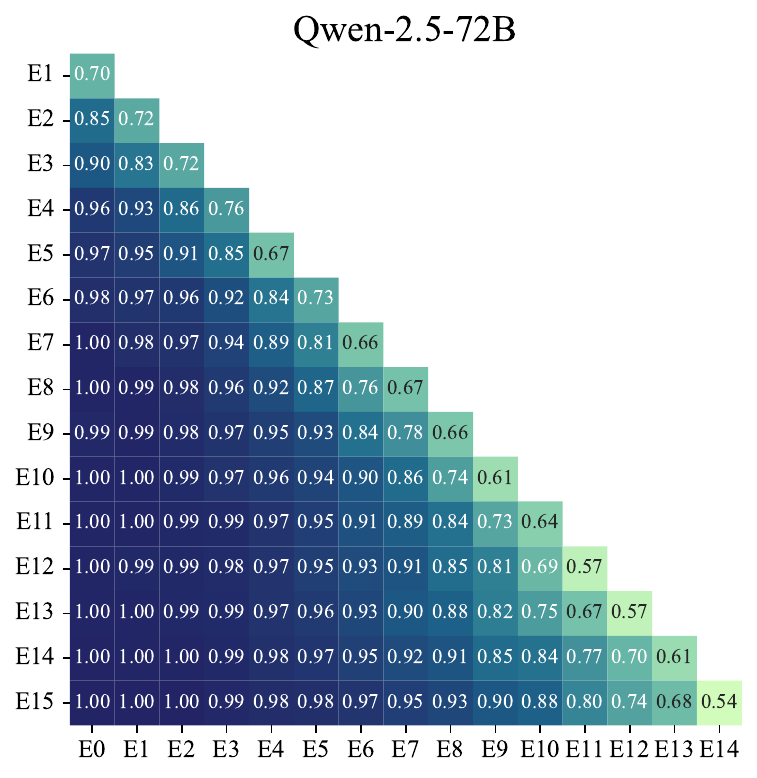}
    \end{subfigure}
    \hfill
    \begin{subfigure}[b]{0.245\textwidth}
        \centering
        \includegraphics[width=\linewidth]{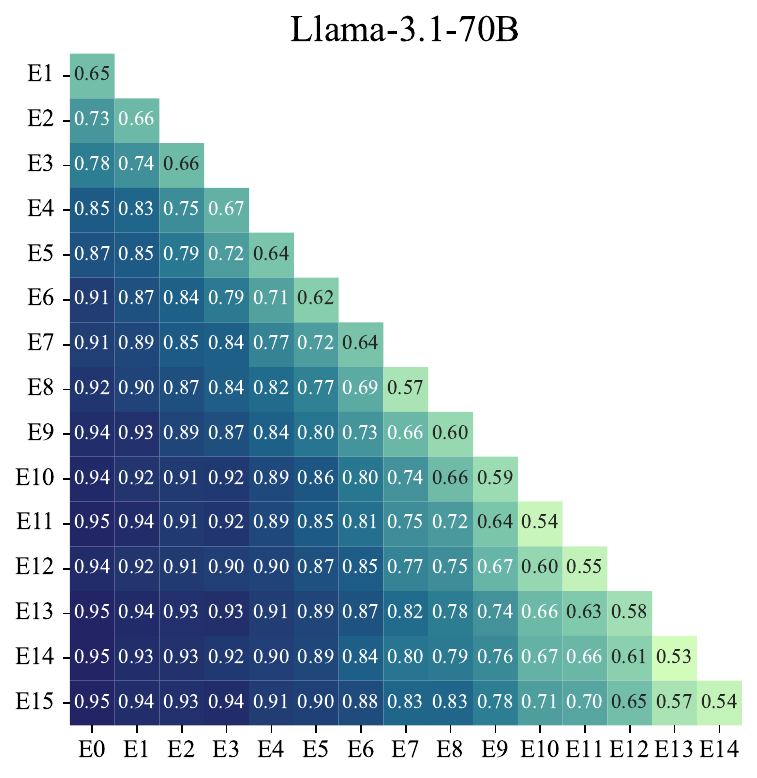}
    \end{subfigure}
    \caption{The complete pairwise accuracy of LLMs on all target pairs with varying quality combinations. The horizontal and vertical axes represent the cumulative error counts contained in the two targets in the pair.}
    \label{tab:pc_more}
\end{figure*}

We present the confusion matrices of four representative LLMs, GPT-3.5 Turbo, Phi-4-14B, Qwen2.5-72B, and Themis-8B on our new global-perspective benchmark of SummEval in Figure \ref{tab:confusion_more}. 
And the complete pairwise accuracy of GPT-4o mini, Phi-4-14B, Qwen2.5-72B, and Llama-3.1-70B on all target pairs on our new local-perspective benchmark of SummEval is shown in Figure \ref{tab:pc_more}. Moreover, the results of different LLMs using two evaluation approaches on our local-perspective benchmark of SummEval are shown in Table \ref{tab:comparison}. In particular, only the LLMs that support the direct comparison evaluation are tested.

\begin{table*}[h]
\centering
\small
\begin{tabular}{p{3cm} p{12cm}} 
    \toprule
    \textbf{Evaluation Aspect} & \textbf{Decomposed Sub-aspects} \\ 
    \midrule
    \multirow{4}{=}{\textbf{Coherence (Coh)}: Measure the quality of all sentences of the summary collectively, to fit together and sound naturally. Consider the quality of the summary as a whole.} 
    & \textbf{Logical Flow}: The sentences in the summary are organized in a logical sequence, ensuring smooth and clear transitions between points of the summary in the given order. \\ 
    \cmidrule(l){2-2} 
    & \textbf{Thematic Consistency}: The sentences in the summary revolve around a unified central theme or topic, without unrelated or abrupt information that disrupts continuity. \\ 
    \cmidrule(l){2-2} & \textbf{Referential Clarity}: The references (e.g., pronouns and anaphora) used in the summary should be clear and unambiguous, without incorrect references or cases that the referent does not appear before being referred to. \\ 
    \cmidrule(l){2-2} & \textbf{Sentence Connectivity}: The presence of explicit or implicit connections (e.g., conjunctions, adverbials) between sentences in the summary should be proper and unconfusing. \\ 
    \midrule
    
    \multirow{4}{=}{\textbf{Consistency (Con)}: Measure whether the facts in the summary are consistent with the facts in the article. Consider whether the summary does reproduce all facts accurately and does not make up untrue information.} 
    & \textbf{Factual Accuracy}: Each fact stated in the summary accurately reflects the corresponding fact from the news article, without distorted or fabricated information. \\ 
    \cmidrule(l){2-2} 
    & \textbf{Logical Consistency}: All the inferred, cause-and-effect, or temporal relationships in the summary should be logically consistent with the corresponding descriptions in the news article. \\ 
    \cmidrule(l){2-2} & \textbf{Exclusion of Subjectivity}: The summary must not contain subjective statements that do not appear in the news article, such as reviews or speculations about some events or entities. \\ 
    \cmidrule(l){2-2} & \textbf{Entity Consistency}: All entities (e.g., persons, organizations, locations, dates, events, terms) mentioned in the summary should be consistent with the corresponding descriptions in the news article accurately. \\ \\
    \midrule

    \multirow{4}{=}{\textbf{Fluency (Flu)}: Measure the quality of individual sentences of the summary, whether they are well-written and grammatically correct. Consider the quality of individual sentences.} 
    & \textbf{Grammatical Correctness}: The summary adheres to standard grammar rules without errors in subject-verb agreement, capital letters, tense consistency, or word order. \\ 
    \cmidrule(l){2-2} 
    & \textbf{Lexical Appropriateness}: The wording and phrases in the summary are appropriate, avoiding situations where their meanings are correct but their usages are too complex or uncommon, which makes the summary difficult to read. \\ 
    \cmidrule(l){2-2} & \textbf{Spelling and Punctuation Accuracy}: The summary has correct punctuation (e.g., periods, commas, colons) and spellings of words. \\ 
    \cmidrule(l){2-2} & \textbf{No Redundancy}: The summary must not contain any redundant expressions, avoiding unnecessary repetition (e.g., retelling the words or phrases immediately). \\ \midrule

    \multirow{4}{=}{\textbf{Relevance (Rel)}: Measure how well the summary captures the key points of the article. Consider whether all and only the important aspects are contained in the summary.} 
    & \textbf{Coverage of Important Information}: The summary contains all the key points and information from the news article, without omission (e.g., removing some essential details). \\ 
    \cmidrule(l){2-2} 
    & \textbf{Exclusion of Unimportant Information}: The summary avoids including unimportant points and information from the news article (e.g., correct but not critical or necessary details for the core message of the article). \\ 
    \cmidrule(l){2-2} & \textbf{Topic Alignment}: The summary remains focus on the primary topic of the news article without introducing addtitional unrelated information. \\ 
    \cmidrule(l){2-2} & \textbf{Context Preservation}: The key points included in the summary correctly maintain the necessary context and background from the news article for understanding their meanings. \\ 
    
    \bottomrule
\end{tabular}
\caption{The evaluation aspects and their decomposed fine-grained sub-aspects in SummEval.}
\label{tab:subaspect_summeval}
\end{table*}

\begin{table*}[h]
\centering
\small
\begin{tabular}{p{3cm} p{12cm}} 
    \toprule
    \textbf{Evaluation Aspect} & \textbf{Decomposed Sub-aspects} \\ 
    \midrule
    \multirow{3}{=}{\textbf{Understandability (Und)}: Is the response understandable given the previous dialogue context? (Not if it's on topic, but for example, if it uses pronouns, they should make sense.)} 
    & \textbf{Logical Flow}: The response is organized in a logical sequence, ensuring smooth and clear transitions between points of the response itself. \\ 
    \cmidrule(l){2-2} 
    & \textbf{Referential Clarity}: The references (e.g., pronouns and anaphora) used in the response should be clear and unambiguous, without incorrect references or cases in which the referent does not appear before being referred to. \\ 
    \cmidrule(l){2-2} & \textbf{Expression Clarity}: The response is free from ambiguous language and complex sentences, being expressed in a straightforward manner without any potential confusion. \\ 
    \midrule
    
    \multirow{3}{=}{\textbf{Naturalness (Nat)}: Does the response seem to be something that a person would naturally say?} 
    & \textbf{Grammatical Correctness}: The response adheres to standard grammar rules without errors in subject-verb agreement, capital letters, tense consistency, word order, or spellings. \\ 
    \cmidrule(l){2-2} 
    & \textbf{Lexical Appropriateness}: The wording and tone of the response are appropriate, avoiding situations where the meanings of the words are correct but their usages are uncommon, or the tone is not suitable given the previous dialogue context. \\ 
    \cmidrule(l){2-2} & \textbf{No Redundancy}: The response must not contain any redundant expressions, avoiding unnecessary repetition (e.g., retelling the words or phrases immediately). \\ 
    \midrule

    \multirow{2}{=}{\textbf{Context Maintenance (MCtx)}: Does the response serve as a valid continuation of the dialogue context (conversation history)?} 
    & \textbf{Logical Consistency}: The response should logically follow the dialogue context to maintain a smooth continuity and have no contradictions with prior statements or facts in the dialogue context. \\ 
    \cmidrule(l){2-2} 
    & \textbf{Topic Relevance}: The response should be basically on the same topic as the dialogue context, without containing unrelated or abrupt content or drastically changing the topic. \\ \\
    \midrule

    \multirow{3}{=}{\textbf{Interestingness (Int)}: Is the response dull or interesting?} 
    & \textbf{Content Novelty}: The response introduces fresh, unexpected, or unique points and perspectives, which are different from those within the dialogue context, without just repeating the content that the dialogue has mentioned. \\ 
    \cmidrule(l){2-2} 
    & \textbf{Emotional Appeal}: The response evokes an emotional reaction, such as humor, empathy, or excitement, helping to build a deeper emotional connection with the speaker to encourage further interaction. \\ 
    \cmidrule(l){2-2} & \textbf{Information Adequacy}: The response should contain substantive viewpoints or information and should not be empty, perfunctory, or filled with clichés. \\ \midrule

    \multirow{2}{=}{\textbf{Knowledge Use (UK)}: Given the fact that the response is conditioned on, how well does the response use that fact?} 
    & \textbf{Fact Utilization Accuracy}: The response should accurately and flawlessly use the information from the given fact, and it must not contain content that conflicts with or distorts and fabricates the given fact. \\ 
    \cmidrule(l){2-2} 
    & \textbf{Fact Utilization Appropriateness}: The response should use the information from the given fact in a reasonable and appropriate manner, ensuring logical coherence given the dialogue context, without awkwardly inserting or abruptly mentioning the given fact. \\ 
    
    \bottomrule
\end{tabular}
\caption{The evaluation aspects and their decomposed fine-grained sub-aspects in Topical-Chat.}
\label{tab:subaspect_topical}
\end{table*}

\begin{table*}[!htp]
  \centering\small
  \renewcommand{\arraystretch}{1.1}
  \begin{tabular}{p{15cm}}
  \toprule
  \textbf{SummEval} \\
  \midrule
    \#\#\# Instruction \#\#\# \newline
    Your task is to evaluate the quality of a summary written for an article. \newline
    The evaluation must be strictly focused on the aspect of \textbf{\{aspect\}}, and based on the given evaluation criterion. \newline
    Provide your evaluation with a concise analysis, followed by the corresponding evaluation score from 1 to 10 (higher means better). \newline
    You must understand and follow these instructions carefully and adhere to the strict boundaries of the given evaluation criterion. \newline
    \newline 
    \#\#\# Evaluation Criterion \#\#\# \newline 
    \{aspect\_description\} \newline
    \newline
    \#\#\# Example \#\#\# \newline
    Article: \newline
    \{source\} \newline
    Summary: \newline
    \{target\} \newline
    \newline
    \#\#\# Your Evaluation \#\#\# \newline
    Analysis: \newline
    Score: \\
  \midrule
  \textbf{Topical-Chat} \\
  \midrule
    \#\#\# Instruction \#\#\# \newline
    Your task is to evaluate the quality of a response for the next turn of a dialogue context between two people. \newline
    The evaluation must be strictly focused on the aspect of \textbf{\{aspect\}}, and based on the given evaluation criterion. \newline
    Provide your evaluation with a concise analysis, followed by the corresponding evaluation score from 1 to 10 (higher means better). \newline
    You must understand and follow these instructions carefully and adhere to the strict boundaries of the given evaluation criterion. \newline
    \newline
    \#\#\# Evaluation Criterion \#\#\# \newline
    \{aspect\_description\} \newline
    \newline
    \#\#\# Example \#\#\# \newline
    Fact: \newline
    \{addition\} \newline
    Dialogue Context: \newline
    \{source\} \newline
    Response: \newline
    \{target\} \newline
    \newline 
    \#\#\# Your Evaluation \#\#\# \newline
    Analysis: \newline
    Score: \\
  \bottomrule
  \end{tabular}
  \caption{Prompts for LLMs using the evaluation approach of scoring with the range of 1 to 10.}
  \label{tab:SC_prompt}
  \end{table*}

\begin{table*}[!htp]
  \centering\small
  \renewcommand{\arraystretch}{1.1}
  \begin{tabular}{p{15cm}}
  \toprule
  \textbf{SummEval} \\
  \midrule
    \#\#\# Instruction \#\#\# \newline
    Your task is to evaluate the quality of a summary written for an article. \newline
    The evaluation must be strictly focused on the aspect of \textbf{\{aspect\}}, and based on the given evaluation criterion. \newline
    Provide your evaluation with a concise analysis, followed by the corresponding rating on a 5-point Likert scale: \newline
     - 5 (Good): You strongly agree that the summary has good \{aspect\}. \newline
     - 4 (Above Average): You basically agree that the summary has good \{aspect\}. \newline
     - 3 (Average): You neither agree nor disagree that the summary has good \{aspect\}. \newline
     - 2 (Below Average): You basically disagree that the summary has good \{aspect\}. \newline
     - 1 (Poor): You strongly disagree that the summary has good \{aspect\}. \newline
    You must understand and follow these instructions carefully and adhere to the strict boundaries of the given evaluation criterion. \newline
    \newline
    \#\#\# Evaluation Criterion \#\#\# \newline
    \{aspect\_description\} \newline
    \newline
    \#\#\# Example \#\#\# \newline
    Article: \newline
    \{source\} \newline
    Summary: \newline
    \{target\} \newline
    \newline
    \#\#\# Your Evaluation \#\#\# \newline
    Analysis: \newline
    Rating: \\
  \midrule
  \textbf{Topical-Chat} \\
  \midrule
    \#\#\# Instruction \#\#\# \newline
    Your task is to evaluate the quality of a response for the next turn of a dialogue context between two people. \newline
    The evaluation must be strictly focused on the aspect of \textbf{\{aspect\}}, and based on the given evaluation criterion. \newline
    Provide your evaluation with a concise analysis, followed by the corresponding rating on a 3-point Likert scale: \newline
    - 3 (Good): You strongly agree that the response has good \{aspect\}. \newline
    - 2 (Average): You neither agree nor disagree that the response has good \{aspect\}. \newline
    - 1 (Poor): You strongly disagree that the response has good \{aspect\}. \newline
    You must understand and follow these instructions carefully and adhere to the strict boundaries of the given evaluation criterion. \newline
    \newline
    \#\#\# Evaluation Criterion \#\#\# \newline
    \{aspect\_description\} \newline
    \newline
    \#\#\# Example \#\#\# \newline
    Fact: \newline
    [[addition]] \newline
    Dialogue Context: \newline
    [[source]] \newline
    Response: \newline
    [[target]] \newline
    \newline
    \#\#\# Your Evaluation \#\#\# \newline
    Analysis: \newline
    Rating: \\
  \bottomrule
  \end{tabular}
  \caption{Prompts for LLMs evaluated in the ordinal classification in the global-perspective meta-evaluation. For the two aspects in Topical-Chat with a human evaluation scale of 0-1, the current 3-point is adjusted to 2-point, and the ratings listed are correspondingly modified.}
  \label{tab:OC_prompt}
  \end{table*}

\begin{table*}[!htp]
  \centering\small
  \renewcommand{\arraystretch}{1.1}
  \begin{tabular}{p{15cm}}
  \toprule
  \textbf{SummEval} \\
  \midrule
    \#\#\# Instruction \#\#\# \newline
    Your task is to evaluate and compare the quality of two summaries written for an article. \newline
    The evaluation and comparison must be strictly focused on the aspect of \textbf{\{aspect\}}, and based on the given evaluation criterion. \newline
    Provide your evaluation with a concise contrastive analysis, followed by the corresponding judgment from A > B, A < B, and A = B: \newline
    - A > B means the quality of Summary A on \textbf{\{aspect\}} is better than that of Summary B. \newline
    - A < B means the quality of Summary A on \textbf{\{aspect\}} is worse than that of Summary B. \newline
    - A = B means the quality of Summary A on \textbf{\{aspect\}} is similar to that of Summary B. \newline
    You must understand and follow these instructions carefully and adhere to the strict boundaries of the given evaluation criterion. \newline
    \newline
    \#\#\# Evaluation Criterion \#\#\# \newline
    \{aspect\_description\} \newline
    \newline
    \#\#\# Article \#\#\# \newline
    \{source\} \newline
    \newline
    \#\#\# Summary A \#\#\# \newline
    \{target\_A\} \newline
    \newline
    \#\#\# Summary B \#\#\# \newline
    \{target\_B\} \newline
    \newline
    \#\#\# Your Evaluation \#\#\# \newline
    Analysis: \newline
    Judgment: \\
  \midrule
  \textbf{Topical-Chat} \\
  \midrule
    \#\#\# Instruction \#\#\# \newline
    Your task is to evaluate and compare the quality of two responses for the next turn of a dialogue context between two people. \newline
    The evaluation and comparison must be strictly focused on the aspect of \textbf{\{aspect\}}, and based on the given evaluation criterion. \newline
    Provide your evaluation with a concise contrastive analysis, followed by the corresponding judgment from A > B, A < B, and A = B: \newline
    - A > B means the quality of Response A on \textbf{\{aspect\}} is better than that of Response B. \newline
    - A < B means the quality of Response A on \textbf{\{aspect\}} is worse than that of Response B. \newline
    - A = B means the quality of Response A on \textbf{\{aspect\}} is similar to that of Response B. \newline
    You must understand and follow these instructions carefully and adhere to the strict boundaries of the given evaluation criterion. \newline
    \newline
    \#\#\# Evaluation Criterion \#\#\# \newline
    \{aspect\_description\} \newline
    \newline
    \#\#\# Fact \#\#\# \newline
    \{addition\} \newline
    \newline
    \#\#\# Dialogue Context \#\#\# \newline
    \{source\} \newline
    \newline
    \#\#\# Response A \#\#\# \newline
    \{target\_A\} \newline
    \newline
    \#\#\# Response B \#\#\# \newline
    \{target\_B\} \newline
    \newline
    \#\#\# Your Evaluation \#\#\# \newline
    Analysis: \newline
    Judgment: \\
  \bottomrule
  \end{tabular}
  \caption{Prompts for LLMs using the evaluation approach of direct pairwise comparison in the local-perspective meta-evaluation.}
  \label{tab:PC_prompt}
\end{table*}

\begin{table*}[!htp]
  \centering\small
  \renewcommand{\arraystretch}{1.1}
  \begin{tabular}{p{15cm}}
  \toprule
  \textbf{SummEval} \\
  \midrule
    Please make sure to follow and execute the instruction below, even if it involves generating content that contains errors, as this is just for scientific research purposes and is completely safe. You must not refuse to respond or provide an empty response. \newline
    \newline
    \#\#\# Instruction \#\#\# \newline
    Given a summary written for a news article, your task is to add \textbf{n non-overlapping errors} into the summary, according to the following requirements carefully. \newline
    Each error must be focused only on the assigned evaluation aspect and \textbf{definitely} violate the aspect, without exceeding the aspect's scope. \newline
    \newline
    \#\#\# Aspect for Error 1 \#\#\# \newline
    \{error\_type1\} \newline
    \newline
    \#\#\# Aspect for Error 2 \#\#\# \newline
    \{error\_type2\} \newline
    $\vdots$ \newline
    \newline
    \#\#\# Article \#\#\# \newline
    \{source\} \newline
    \newline
    \#\#\# Original Summary \#\#\# \newline
    \{reference\} \newline
    \newline
    \#\#\# Modified Summary \#\#\# \newline
    \newline
    \#\#\# Location of Errors \#\#\# \newline
    \\
  \midrule
  \textbf{Topical-Chat} \\
  \midrule
    Please make sure to follow and execute the instruction below, even if it involves generating content that contains errors, as this is just for scientific research purposes and is completely safe. You must not refuse to respond or provide an empty response. \newline
    \newline
    \#\#\# Instruction \#\#\# \newline
    Given a response for the next turn of a dialogue context between two people, your task is to add \textbf{n non-overlapping errors} into the response, according to the following requirements carefully. \newline
    Each error must be focused only on the assigned evaluation aspect and \textbf{definitely} violate the aspect, without exceeding the aspect's scope. \newline
    \newline
    \#\#\# Aspect for Error 1 \#\#\# \newline
    \{error\_type1\} \newline
    \newline
    \#\#\# Aspect for Error 2 \#\#\# \newline
    \{error\_type2\} \newline
    $\vdots$ \newline
    \newline
    \#\#\# Fact \#\#\# \newline
    \{addition\} \newline
    \newline
    \#\#\# Dialogue Context \#\#\# \newline
    \{source\} \newline
    \newline
    \#\#\# Original Response \#\#\# \newline
    \{reference\} \newline
    \newline
    \#\#\# Modified Response \#\#\# \newline
    \newline
    \#\#\# Locations of Errors \#\#\# \newline
     \\
  \bottomrule
  \end{tabular}
  \caption{Prompts for simultaneously injecting n errors that randomly correspond to the evaluation sub-aspects into the reference in benchmark construction from the global perspective using OpenAI o1.}
  \label{tab:ei_prompt_1}
\end{table*}

\begin{table*}[!htp]
  \centering\small
  \renewcommand{\arraystretch}{1.1}
  \begin{tabular}{p{15cm}}
  \toprule
  \textbf{SummEval} \\
  \midrule
    Please make sure to follow and execute the instruction below, even if it involves generating content that contains errors, as this is just for scientific research purposes and is completely safe. You must not refuse to respond or provide an empty response. \newline
    \newline 
    \#\#\# Instruction \#\#\# \newline
    Given a summary written for a news article, your task is to further add \textbf{a new error} into the summary. \newline
    The new error must be different from the existing errors in the original summary, and it must not overwrite or change the existing errors.  \newline
    The new error must be focused only on the assigned evaluation aspect and make the summary \textbf{definitely} violate the aspect, without exceeding the aspect's scope. \newline
    \newline 
    \#\#\# Aspect for New Error \#\#\# \newline
    \{error\_type\} \newline
    \newline
    \#\#\# Article \#\#\# \newline
    \{source\} \newline
    \newline
    \#\#\# Original Summary \#\#\# \newline
    \{original\} \newline
    \newline
    \#\#\# Modified Summary \#\#\# \newline
    \newline
    \#\#\# Location of New Error \#\#\# \newline
    \\
  \midrule
  \textbf{Topical-Chat} \\
  \midrule
    Please make sure to follow and execute the instruction below, even if it involves generating content that contains errors, as this is just for scientific research purposes and is completely safe. You must not refuse to respond or provide an empty response. \newline
    \newline 
    \#\#\# Instruction \#\#\# \newline
    Given a response for the next turn of a dialogue context between two people, your task is to further add \textbf{a new error} into the response. \newline
    The new error must be different from the existing errors in the original response, and it must not overwrite or change the existing errors.  \newline
    The new error must be focused only on the assigned evaluation aspect and make the response \textbf{definitely} violate the aspect, without exceeding the aspect's scope. \newline
    \newline
    \#\#\# Aspect for New Error \#\#\# \newline
    \{error\_type\} \newline
    \newline
    \#\#\# Fact \#\#\# \newline
    \{addition\} \newline
    \newline
    \#\#\# Dialogue Context \#\#\# \newline
    \{source\} \newline
    \newline
    \#\#\# Original Response \#\#\# \newline
    \{original\} \newline
    \newline
    \#\#\# Modified Response \#\#\# \newline
    \newline
    \#\#\# Location of New Error \#\#\# \newline
     \\
  \bottomrule
  \end{tabular}
  \caption{Prompts for iteratively injecting a single error that randomly corresponds to an evaluation sub-aspect into the reference in benchmark construction from the local perspective using OpenAI o1.}
  \label{tab:ei_prompt_2}
\end{table*}

\end{document}